\documentclass[10pt,twocolumn,letterpaper]{article}

\usepackage{cvpr}  
\usepackage[dvipsnames]{xcolor}
\usepackage{tikz}

\newcommand{\OURS}{FaceTalk}
\newcommand{\dataTotal}{1000}

\renewcommand{\paragraph}[1]{\smallskip\noindent\textbf{#1}}

\usepackage{graphicx}
\usepackage{amsmath}
\usepackage{amssymb}
\usepackage{booktabs}
\usepackage{comment}
\usepackage[accsupp]{axessibility}

\definecolor{cvprblue}{rgb}{0.21,0.49,0.74}
\usepackage[pagebackref,breaklinks,colorlinks,citecolor=cvprblue]{hyperref}

\usepackage[capitalize]{cleveref}
\crefname{section}{Sec.}{Secs.}
\Crefname{section}{Section}{Sections}
\Crefname{table}{Table}{Tables}
\crefname{table}{Tab.}{Tabs.}

\begin{document}

\title{\OURS: Audio-Driven Motion Diffusion for Neural Parametric Head Models}

\author{Shivangi Aneja\textsuperscript{\rm 1}
$\quad$
Justus Thies\textsuperscript{\rm 2,}\textsuperscript{\rm 3}
$\quad$
Angela Dai\textsuperscript{\rm 1}
$\quad$
Matthias Nie\ss ner\textsuperscript{\rm 1} \vspace{0.3em}\\
{\normalsize \textsuperscript{\rm 1}Technical University of Munich} \quad
{\normalsize \textsuperscript{\rm 2}MPI-IS, T{\"u}bingen} \quad
{\normalsize \textsuperscript{\rm 3}TU Darmstadt} \quad
}

\affiliation{
    \textsuperscript{\rm 1}Institution1\\
    {\tt\small {email1@example.com}}
}

\twocolumn[{%
\renewcommand\twocolumn[1][]{#1}%
\maketitle
\begin{center}
    \centering
    \includegraphics[width=1.0\linewidth]{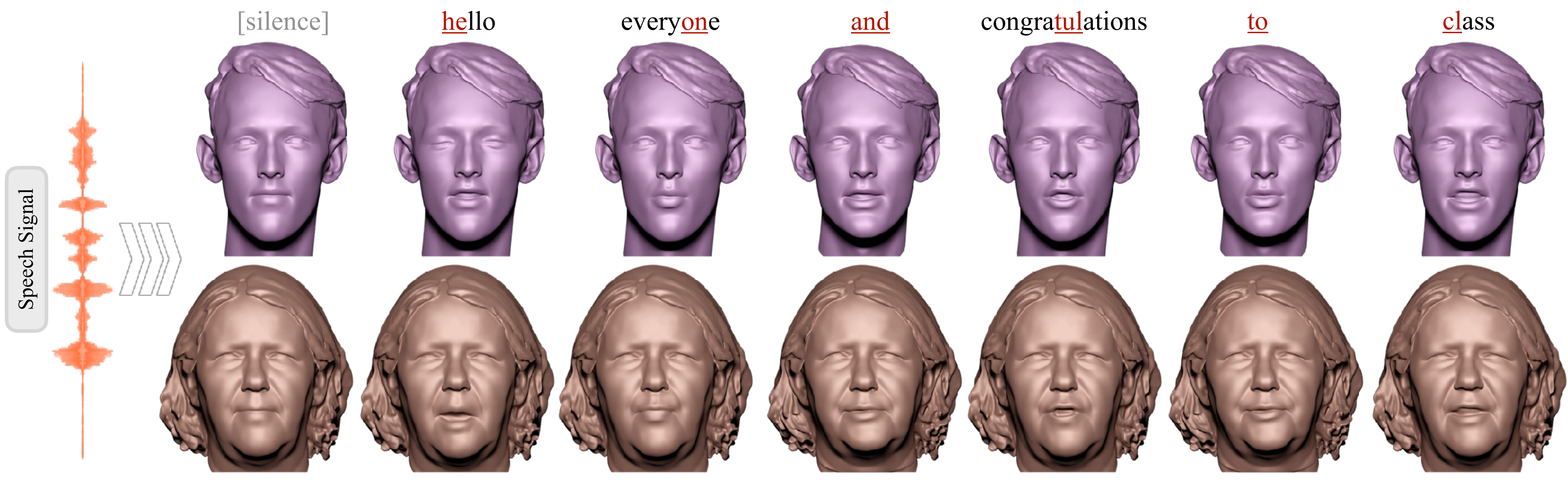}
    \captionof{figure}{
    Given an input speech signal, we propose a diffusion-based approach to synthesize high-quality and temporally consistent 3D motion sequences of high-fidelity human heads as neural parametric head models. Our method can generate a diverse set of expressions (including wrinkles and eye blinks) and the generated mouth motion is temporally synchronized with the given audio signal.
    }
    \label{fig:teaser}
\end{center}%
}]

\begin{abstract}

We introduce \OURS{}\footnote{Project Page: \url{https://shivangi-aneja.github.io/projects/facetalk}}, a novel generative approach designed for synthesizing high-fidelity 3D motion sequences of talking human heads from input audio signal.
To capture the expressive, detailed nature of human heads, including hair, ears, and finer-scale eye movements, we propose to couple speech signal with the latent space of neural parametric head models to create high-fidelity, temporally coherent motion sequences. 
We propose a new latent diffusion model for this task, operating in the expression space of neural parametric head models, to synthesize audio-driven realistic head sequences.
In the absence of a dataset with corresponding NPHM expressions to audio, we optimize for these correspondences to produce a dataset of temporally-optimized NPHM expressions fit to audio-video recordings of people talking.
To the best of our knowledge, this is the first work to propose a generative approach for realistic and high-quality motion synthesis of volumetric human heads, representing a significant advancement in the field of audio-driven 3D animation. Notably, our approach stands out in its ability to generate plausible motion sequences that can produce high-fidelity head animation coupled with the NPHM shape space. Our experimental results substantiate the effectiveness of \OURS{}, consistently achieving superior and visually natural motion, encompassing diverse facial expressions and styles, outperforming existing methods by 75\% in perceptual user study evaluation.
\end{abstract}

\section{Introduction}
\label{sec:intro}

Modeling 3D animation of humans has a wide range of applications in the realm of digital media, including animated movies, computer games, and virtual assistants. In recent years, there have been numerous works proposing generative approaches for motion synthesis of human bodies, enabling the animation of human skeletons conditioned on various signals such as action~\cite{action2motion2020, petrovich21actor, TEACH_3DV_2022}, language~\cite{ahuja2019language2pose, petrovich22temos, tevet2023human, zhang2023generating, kim2022flame, SINC_ICCV_2023} and music~\cite{tseng2022edge, alexanderson2023listen}. 
While human faces are critical to synthesis of humans, generative synthesis of 3D faces in motion has focused on 3D morphable models (3DMMs) leveraging linear blendshapes \cite{blanz2023morphable,flame2017} to represent head motion and expression. 
Such models characterize a disentangled space of head shape and motion, but lack the capacity to comprehensively represent the complexity and fine-grained details of human face geometry in motion (e.g., hair, skin furrowing during motion, etc.).

Thus, we propose to represent animated head sequences with a volumetric 3D head representation, leveraging the expressive representation space of neural parametric head models (NPHMs)~\cite{giebenhain2023nphm,giebenhain2024mononphm}.
NPHMs offer a flexible representation capable of handling complex and irregular facial expressions (e.g., blinking, skin creasing), along with a high-fidelity shape space including the head, hair, and ears, making them a much more suitable choice for face animation.
We address the challenging task of creating an audio-conditional generative animation model for this volumetric representation.

We design the first transformer-based latent diffusion model for audio-driven head animation synthesis. 
Our diffusion model operates in the latent NPHM expression space to generate temporally coherent expressions consistent with an input audio signal represented with Wave2Vec 2.0~\cite{baevski2020wav2vec}. 
In the absence of paired audio-NPHM data, we optimize for corresponding NPHM expressions to fit to multi-view video recordings of people speaking, generating train supervision for our task.
As NPHMs are designed for static (frame-by-frame) expressions without temporal consistency, we employ both geometric and temporal priors to produce temporally consistent optimized motion sequences.
This enables training our audio-head diffusion model to synthesize realistic speech-conditioned 3D head sequences, which are capable of capturing high-frequency details like wrinkles and eye blinks present in the face region. 
Our method takes the first step towards simplifying the task of high-fidelity facial motion generation of 3D faces for content creation applications. 

In summary, we propose the first latent diffusion model for the creation of audio-conditioned animations of volumetric avatars.
By producing volumetric head animation, our generative model is highly expressive yet efficient compared to existing 3D methods. We also demonstrate control over motion style, using classifier-free guidance to adjust the strength of the stylistic expression.
\begin{figure*}[t!]
    \centering
    \includegraphics[width=1.0\linewidth]{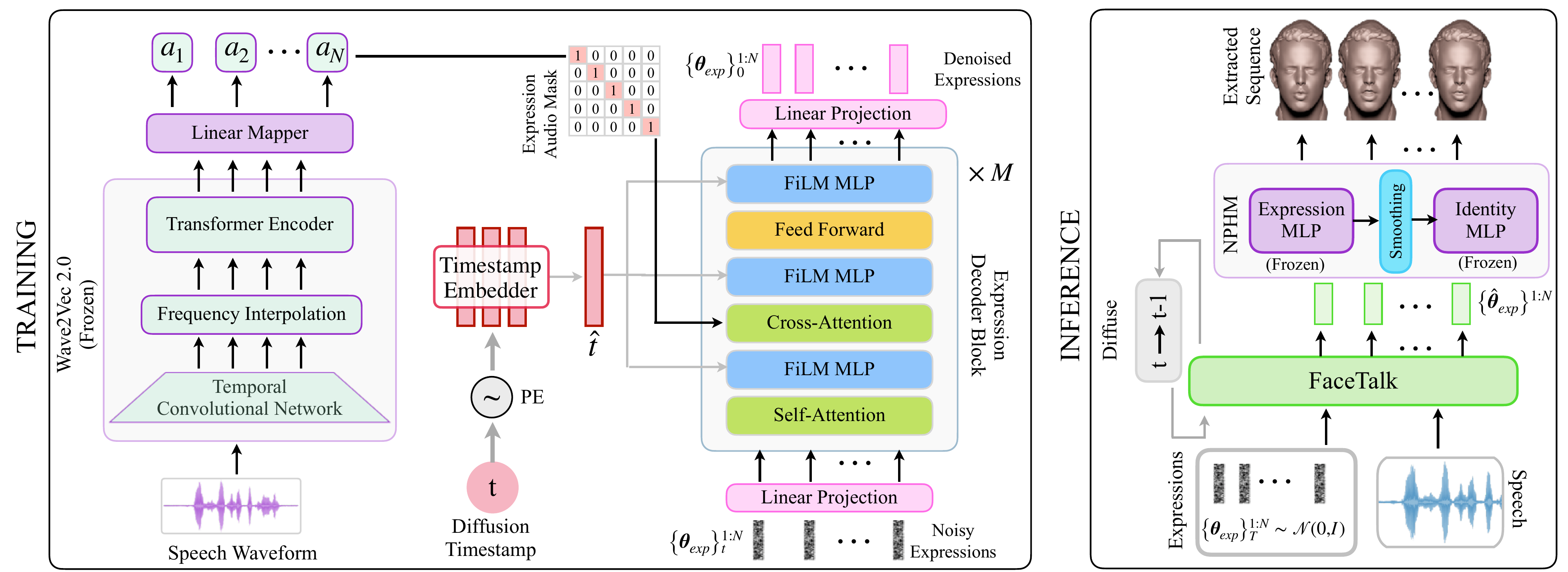}
    \caption{\textbf{Pipeline Overview.} 
    \OURS{} uses frozen Wave2Vec 2.0~\cite{baevski2020wav2vec} to extract audio embeddings from a speech signal. The diffusion timestamp is embedded using a timestamp embedder. The expression decoder employs a multi-head transformer decoder~\cite{vaswani2023attention} with FiLM~\cite{perez2017film} layers, interleaved between Self-Attention, Cross Attention, and FeedForward layers, to incorporate diffusion timestamp. During training, the model is trained to denoise the noisy expression sequences from timestamp $t$. At inference, \OURS{} denoises the gaussian noise sequence $ \big\{ \boldsymbol{\theta}_{exp} \big\}_{T}^{1:N} \sim \mathcal{N}(0,\boldsymbol{{I}})$ iteratively until $t=0$, yielding the estimated final sequence $ \big\{ \hat{\boldsymbol{\theta}}_{exp} \big\}^{1:N}$. These are then input to the frozen NPHM model, utilizing facial smoothing, and mesh sequences are extracted using MC~\cite{marching_cubes}.
    }
    \label{fig:method_diffusion}
\end{figure*}

\section{Related Work}
\label{sec:related_work}

\textbf{Facial Animation.} Our proposed method is the first work for audio-conditioned latent diffusion of volumetric head avatars.
There is a large corpus of research works in the field of 2D audio-driven facial animation operating on RGB videos, synthesizing 2D sequences directly~\cite{chung2017said, Wiles18, chen2018lip, suwajanakorn2017obama, wave2lip2020, chen2019hierarchical, lipgan2019, make_it_talk2020, vougioukas2018endtoend, vougioukas2019realistic, zhou2018talking, cascade2020, chen2020talking, emotion22, gururani2022space, shen2023difftalk, stypulkowski2022diffused}. However, these methods operate in pixel space and do not produce any geometric information. Another line of work also operating on RGB videos but using intermediate 3D representations are based on 3DMMs~\cite{song2020everybodys, thies2020nvp, ji2021audio-driven, Doukas_2021_ICCV, tang2022memories}. Although these methods generate RGB videos in the end, they use 3DMMs which produces very sparse geometric information.
%

Recent works based on radiance fields~\cite{guo2021adnerf, yao2022dfa, shen2022dfrf, liu2022semantic, Gafni_2021_CVPR, kirschstein2023nersemble} have also gained popularity due to their capability to model densities directly from images. These methods generate impressive RGB videos but the underlying geometry learned is highly imperfect (no segregation between background and facial region) as shown by Chan \etal~\cite{Chan2021}. Additionally, these methods require identity-specific training, and thus can not be used for content creation applications.

Learning animation of 3D meshes directly is much more promising, but only a handful of methods exist~\cite{karras2017audio, VOCA2019, richard2021meshtalk, faceformer2022,  xing2023codetalker, Thambiraja_2023_ICCV, Peng_2023_ICCV, EMOTE}. A vast majority of these works model speech-conditioned animation for an artist-designed template mesh. Although these methods can match the facial motion with the speech signal, one limitation of these methods is their incapability to represent fine-scale details present in faces. 
Another downside of these methods is that these approaches learn a deterministic model producing no/muted motion in the upper region of the face, thus limiting them from being able to produce realistic motion. In this work, we solve these issues by proposing a generative model that can operate in the compact and detailed latent space of neural parametric head models, thus capable of representing fine-scale facial details and synthesizing a diverse set of expressions and speaking styles.

\textbf{Diffusion Models for Generative Synthesis.}
In recent years, diffusion models have experienced a surge of interest as highly expressive and efficient generative models. These models have demonstrated strong performances as a generative model for a variety of domains such as images~\cite{ho2020denoising, dhariwal2021diffusion, rombach2022highresolution, ramesh2022hierarchical, saharia2022photorealistic, zhang2023adding}, videos~\cite{fan2023hierarchical, voleti2022MCVD, blattmann2023videoldm, dhesikan2023sketching}, speech~\cite{huang2022fastdiff, kong2021diffwave, lam2022bddm} and motion~\cite{zhu2023taming, azadi2023makeananimation, tevet2023human, kim2022flame, tseng2022edge, alexanderson2023listen}. 

2D facial animation has also seen some progress with diffusion models~\cite{shen2023difftalk, stypulkowski2022diffused}. DiffusedHeads~\cite{stypulkowski2022diffused} operates in pixel space, making the sampling process very slow. Although DiffTalk~\cite{shen2023difftalk} learns the diffusion process in the latent space, the sampling is performed in an autoregressive fashion, limiting sampling efficiency. Moreover, these methods operate on RGB videos, and though they achieve high-quality results on a per-frame basis, consistency and coherence across different timesteps remain challenging due to temporal jitter. Hence, these methods can not be applied directly to the parametric head models. Concurrent to our work, DiffPoseTalk~\cite{sun2023diffposetalk} and FaceDiffuser~\cite{facediffuser} propose a diffusion-based approach for animating 3D meshes from speech signal. However, these methods require additional conditions like style embedding, generate sequences in an autoregressive fashion coupled with diffusion denoising making them very slow and still they can not produce fine-scale facial details due to the use of 3DMMs which are limited in expressivity.
In contrast, we synthesize the entire sequence simultaneously making it much faster and operate directly in the latent space of highly expressive NPHMs, producing high-fidelity and temporally consistent results.

\section{Preliminaries}
\label{sec:prelimaries}

\paragraph{NPHM.}
In contrast to traditional 3D morphable models, which remain limited in expressive detail, we employ the Neural Parametric Head Model (NPHM) representation~\cite{giebenhain2023nphm,giebenhain2024mononphm}.
Similar to traditional 3DMMs, NPHM disentangles a human face into an identity and expression space; however, the shape and expression space are volumetric, enabling expressive capture of details such as fine-scale eye movements, hair, ears, etc.
NPHM uses two auto-decoder style neural networks (a) Identity Network $\big\{\mathcal{I} \big\}$ to represent the overall facial shape and (b) Expression Network \big\{$\mathcal{E} \big\}$ to represent the facial movements such as jaw pose, wrinkles, eyeblinks, etc., jointly denoted as $\mathcal{F}  = \big\{\mathcal{I}, \mathcal{E} \big\}$. The identity and expressions latent codes $\big\{ \boldsymbol{{\theta}_{id}}, \boldsymbol{{\theta}_{exp}} \big\}$ define the facial shape and expression respectively. Unlike 3DMMs which predict a fixed-topology mesh that can not model fine surface details, NPHM can handle different hairstyles, wrinkles and complex facial expressions. NPHM represents identities with a signed distance field (SDF) in canonical space and models the expressions as deformations onto the facial region. Mathematically, it can be defined as:
\begin{equation}
    \mathcal{F(\boldsymbol{x_i}, \boldsymbol{{\theta}_{id}},  \boldsymbol{{\theta}_{exp}}) \rightarrow \boldsymbol{s_i}}: \mathbb{R}^{3} \times \mathbb{R}^{\boldsymbol{{\theta}_{id}}}\times \mathbb{R}^{\boldsymbol{{\theta}_{exp}}}  \rightarrow \mathbb{R},
\end{equation}
where $\boldsymbol{x_i} \in \mathbb{R}^{3}$ represents the query points, $\boldsymbol{{\theta}_{id}} \in \mathbb{R}^{1344}$ represents identity latent code, $\boldsymbol{{\theta}_{exp}} \in \mathbb{R}^{200}$ represents expression latent code and $\boldsymbol{s_i}$ denotes the predicted SDF. The mesh is then extracted as the zero-level iso-surface decision boundary using Marching Cubes~\cite{marching_cubes}.

\paragraph{Diffusion. }
Diffusion models are a class of generative models that consist of a forward and reverse process. The forward process converts the original structured data distribution into Gaussian noise, modeled following a fixed Markov chain as:
\begin{equation}
    q(\boldsymbol{x}_{1:T}|\boldsymbol{x}_0) = \prod_{t=1}^{T} q(\boldsymbol{x}_t|\boldsymbol{x}_{t-1}),
\end{equation}
where $q(\boldsymbol{x}_t|\boldsymbol{x}_{t-1})$  denotes the forward process adding white noise to the original data distribution $\boldsymbol{x}_0$. Mathematically, the forward process can be written as: 
\begin{equation}
\label{eq:noising}
 q(\boldsymbol{x}_t|\boldsymbol{x}_0) \sim   \mathcal{N}(\sqrt{\overline{\alpha}_t} \boldsymbol{x}_0, (1-\overline{\alpha}_t)\boldsymbol{{I}}),
\end{equation}
where $\overline{\alpha}_t \in (0, 1)$ are constants following a decreasing cosine schedule such that when $\overline{\alpha}_t$ tends to 0, we can approximate $\boldsymbol{x}_T \sim \mathcal{N}(0,\boldsymbol{{I}})$. The reverse diffusion process progressively denoises samples from $\boldsymbol{x}_T \sim \mathcal{N}(0,\boldsymbol{{I}})$ into samples from a learned distribution $\boldsymbol{x}_0$. In our experiments, we learn a generative model $\mathcal{G}_{\theta}$ to reverse the forward diffusion process by learning to estimate $ \mathcal{G}_{\theta}(\boldsymbol{x}_t, t, \boldsymbol{c}) = \boldsymbol{\hat{x}} \approx \boldsymbol{x}  $, where $\boldsymbol{x}$ refers to predicting the cleaned samples directly, $\boldsymbol{x}_t$ refers to noisy input at diffusion timestamp $t$. Given a conditional signal $\boldsymbol{c}$, we optimize model parameters $\boldsymbol{\theta}$ for all diffusion timestamps $t$ using the following objective: 

\begin{equation}
\label{eq:diffusion_loss}
    \mathcal{L}_{\boldsymbol{\theta}} = \mathbb{E}_{\boldsymbol{x}, t} \big[ \| \boldsymbol{x} - \mathcal{G}_{\theta}(\boldsymbol{x}_t, t, \boldsymbol{c}) \|_2^2  \big].
\end{equation}

\section{Method}
\label{sec:method}
\OURS{} performs high-fidelity and temporally consistent generative synthesis of motion sequences of heads, conditioned on audio signal. 
In order to characterize complex face motions and fine-scale movements, we synthesize realistic heads in the latent expression space of a parametric model for volumetric head representations, i.e., neural parametric head models (NPHMs) \cite{giebenhain2023nphm}.
We, thus, develop a speech-conditioned latent diffusion model to synthesize temporally coherent head expression sequences that are coupled with the NPHM shape space to produce complex, realistic head animations of different identities.
An overview of our approach is illustrated in Fig.~\ref{fig:method_diffusion}.

\paragraph{Audio Encoding.} We employ a state-of-the-art pre-trained speech model Wave2Vec 2.0~\cite{baevski2020wav2vec} to encode the audio signal. Specifically, we first use the audio feature extractor made up of temporal convolution layers (TCN) to extract audio feature vectors $ \{ a_i \}_{i=1}^{N_a}$ from the raw waveform. This is followed by the Frequency Interpolation layer that aligns the input audio signal $ \{ a_i \}_{i=1}^{N_a}$ (captured at frequency $f_a$ = 16kHz) with our dataset $ \{ a_i \}_{i=1}^{N_e}$ (captures at framerate $f_e$ = 24Hz). Finally, a stacked multi-layer transformer encoder network processes these resampled features and outputs the processed and aligned audio feature vectors. The audio encoder is initialized with the pre-trained wav2vec 2.0 weights, followed by a feedforward layer to project the aligned audio features into the latent space of our expression decoder model. These aligned audio features are denoted as: 

\begin{equation}
    \boldsymbol{A}^{1:N} = \{ \boldsymbol{a}_i \}_{i=1}^{N}.
\end{equation}

These audio features $\boldsymbol{A}^{1:N}$ are then fed to the cross-attention layers of the expression decoder via the expression-audio alignment mask $\mathcal{M}$ to learn speech-conditioned expression features during training. 

\paragraph{Expression Encoding.} We train our expression decoder network on optimized NPHM expression sequences $\big\{ \boldsymbol{\theta}_{exp} \big\}^{1:N} $ (obtained in Section~\ref{sec:dataset_generation}), where $N$ refers to number of frames in a sequence. We train a diffusion-based stacked multi-layer transformer~\cite{vaswani2023attention} decoder network to synthesize facial expressions in the latent space of the NPHM model. During training, following forward diffusion (Eq.~\ref{eq:noising}) we add noise for a randomly sampled diffusion timestamp  $t \sim \mathtt{Uniform}(0, T)$ to create noisy expression codes $\big\{ \boldsymbol{\theta}_{exp} \big\}^{1:N}_{t}$. These noisy expression codes are then projected to the latent space of our model via a linear layer, followed by a stack of transformer decoder blocks, and then projected back to the original NPHM space using another linear layer. To embed diffusion timestamp in the latent space of our model, we apply sinusoidal embedding and process it through a three-layer MLP. Next, to fuse diffusion timestamp into the model, we use a one-layer FiLM (feature-wise linear modulation) network~\cite{perez2017film}  between the multi-head self-attention, multi-head cross-attention and feedforward layers of the transformer decoder block, which is critical for the model to produce high-quality expressions as we show in the results.  We leverage a look-ahead binary target mask $\mathcal{T} \in  \mathbb{R}^{N \times N}$ in the multi-head self-attention layers to prevent the model from peeking into the future expression codes. Mathematically, it can be written as:
\begin{equation}
      \mathcal{T}_{ij} =   
    \begin{cases}
        True \quad \text{if } i \leq j\\
        False \quad \text{else}\\ 
    \end{cases}
\end{equation}
where $\mathcal{T}_{ij}$ refers to the $(i,j)^{th}$ element of the matrix and $1 \leq i,j \leq N $.
The audio features $\boldsymbol{A}^{1:N}$ are fused into the network via the multi-head cross-attention layers. We leverage the expression-audio alignment mask $\mathcal{M}$ to fuse the audio features into the network to ensure that mouth poses learned by the expression codes are consistent with the speech signal.  The binary mask $\mathcal{M} \in  \mathbb{R}^{N \times N}$ is Kronecker delta function $\delta_{ij}$ such that the audio features for $i^{th}$ timestamp attend to expression features at the $j^{th}$ timestamp if and only if $i=j$. Mathematically,

\begin{equation}
      \mathcal{M} =  \delta_{ij} =  
    \begin{cases}
        True \quad \text{if } i=j\\
        False \quad \text{if } i \neq j\\ 
    \end{cases}
\end{equation}

We demonstrate in the results (Section~\ref{sec:results}) that this alignment is crucial for learning audio-consistent expression codes. As a by-product, this further enables the model to generalize to arbitrary long audio signals during inference which we also show in our results. Our model is trained with the diffusion loss in Eq.~\ref{eq:diffusion_loss}, with input $\boldsymbol{x} = \big\{ \boldsymbol{\theta}_{exp} \big\}^{1:N}$ and conditioning $\boldsymbol{c} = \boldsymbol{A}^{1:N}$.

\paragraph{Expression Augmentation.} Unlike other domains (like text-to-motion) where the synthesized motion can vary dramatically, speech-driven 3D facial animation is more constrained, requiring precise mouth alignment with the audio signal, making it prone to overfitting. To alleviate this, we propose a novel augmentation strategy capable of producing diverse expressions for a given audio signal. Our key insight is to augment the dataset to generate different expression codes for the same speech signal by randomly amplifying and suppressing its magnitude. Specifically, we randomly sample modulation factor $r$ within the bounds $[a,b]$ as:
\begin{equation}
    r \sim \mathtt{Uniform}(a, b),
\end{equation}
and scale the expression codes as $ \big\{ r \times \boldsymbol{\theta}_{exp} \big\}^{1:N} $.  We show in the results this augmentation strategy helps the model synthesize diverse expressions.

\paragraph{Sampling.} \OURS{} uses a diffusion-based framework to learn to synthesize NPHM expression sequences of $N$ frames $ \big\{ \boldsymbol{\theta}_{exp} \big\}^{1:N}  \in \mathbb{R}^{N \times 200}$. At each of the denoising timestep t, \OURS{} predicts the denoised sample and noises it back to timestamp ${t-1}$ :
\begin{equation}
    \big\{ \boldsymbol{\theta}_{exp} \big\}^{1:N}_{t-1} = \mathcal{G}_{\theta}\big( \big\{ \boldsymbol{\theta}_{exp} \big\}^{1:N}_{t}, \boldsymbol{A}^{1:N}, t-1  \big),
\end{equation}
terminating when it reaches t = 0. We train our model using classifier-free guidance~\cite{ho2022classifierfree}. We implement classifier-free guidance by randomly replacing the audio with null conditioning $\boldsymbol{A}^{1:N} = \Phi$ during training with 25\% probability. During inference, we can use the weighted combination of conditional and unconditionally generated samples:
\begin{equation}
    \big\{ \boldsymbol{\theta}_{exp}^{c} \big\}^{1:N}_{t} = w . \big\{ \boldsymbol{\theta}_{exp}^{c} \big\}^{1:N}_{t} + (1-w) . \big\{ \boldsymbol{\theta}_{exp}^{u} \big\}^{1:N}_{t},
\end{equation}
where $\theta_{exp}^{c}$ and $\theta_{exp}^{u}$ refers to conditionally and unconditionally generated samples. To amplify the audio conditioning, we can use guidance strength $w > 1$.

\paragraph{Sequence Generation.} The expression codes predicted above $\big\{ \boldsymbol{\hat{\theta}}_{exp} \big\}^{1:N}$ are then passed to the pretrained NPHM expression mapper $ \big\{ \mathcal{E} \big\} $ to obtain expression deformations $\big\{ \boldsymbol{\delta}_{exp} \big\}^{1:N}$. We then apply smoothing based on a Gaussian kernel to these deformations to remove the unwanted wiggle in the head and neck regions and to ensure that generating sequences are free from flickering artifacts. Specifically, we first define a control center $\mathcal{C}$ on the mouth region in the canonical space as:
\begin{equation}
    \mathcal{C} = [c_x, c_y, c_z].
\end{equation}
Next, using a 3D gaussian kernel with standard deviation $ \Sigma = [\sigma_x, \sigma_y, \sigma_z] $, centered at $\mathcal{C}$, for each point sampled from 3D grid $\mathcal{P} \in \mathbb{R}^{|X| \times |Y| \times |Z|}$ as $\mathcal{P}_{xyz} = [p_x, p_y, p_z]$, we compute its distance from $\mathcal{C}$ as:
\begin{equation}
    d_{xyz} = \sqrt{ \frac{(p_x - c_x)^2}{\sigma_x^2} + \frac{(p_y - c_y)^2}{\sigma_y^2} + \frac{(p_z - c_z)^2}{\sigma_z^2}   }.
\end{equation}
Based on this distance, smoothing weights are obtained from the Gaussian kernel with min-max normalization as:
\begin{equation}
    w_{xyz} = \frac{1}{2 \pi \sigma_x \sigma_y \sigma_z } . e^{- (\frac{1}{2} \times d_{xyz}^2)}
\end{equation}
\begin{equation}
    w_{xyz}^{k} = \frac{w_{xyz}^{k} - \text{min}\big(w_{xyz}^{1:M}\big) }{\text{max}\big(w_{xyz}^{1:M}\big) - \text{min}\big(w_{xyz}^{1:M}\big)},
\end{equation}
where $M=|X| \times |Y| \times |Z|$. The obtained expression deformations $\big\{ \boldsymbol{\delta}_{exp} \big\}^{1:N}$ are then multiplied with these smoothing weights $\big\{ \boldsymbol{w} \big\}^{1:N} = \big\{ (w^1, w^2, \cdots w^M)\big\}^{1:N} $ to obtain smoothed expression deformations as:
\begin{equation}
    \big\{ \boldsymbol{\delta}_{exp}^{s}\big\}^{1:N} = \big\{ \boldsymbol{\delta}_{exp}\big\}^{1:N} \times  \big\{ \boldsymbol{w} \big\}^{1:N}.
\end{equation}
These smoothed expression deformations $\big\{ \boldsymbol{\delta}_{exp}^{s}\big\}^{1:N}$, along with identity $\boldsymbol{\theta_{id}}$ and predicted expression codes $\big\{ \boldsymbol{\hat{\theta}}_{exp} \big\}^{1:N}$ are passed to the NPHM identity MLP $\big\{ I \big\}$ to obtain smoothed SDF, from which meshes are extracted using MC~\cite{marching_cubes}.

\begin{figure}[t!]
    \centering
    \includegraphics[width=1.0\linewidth]{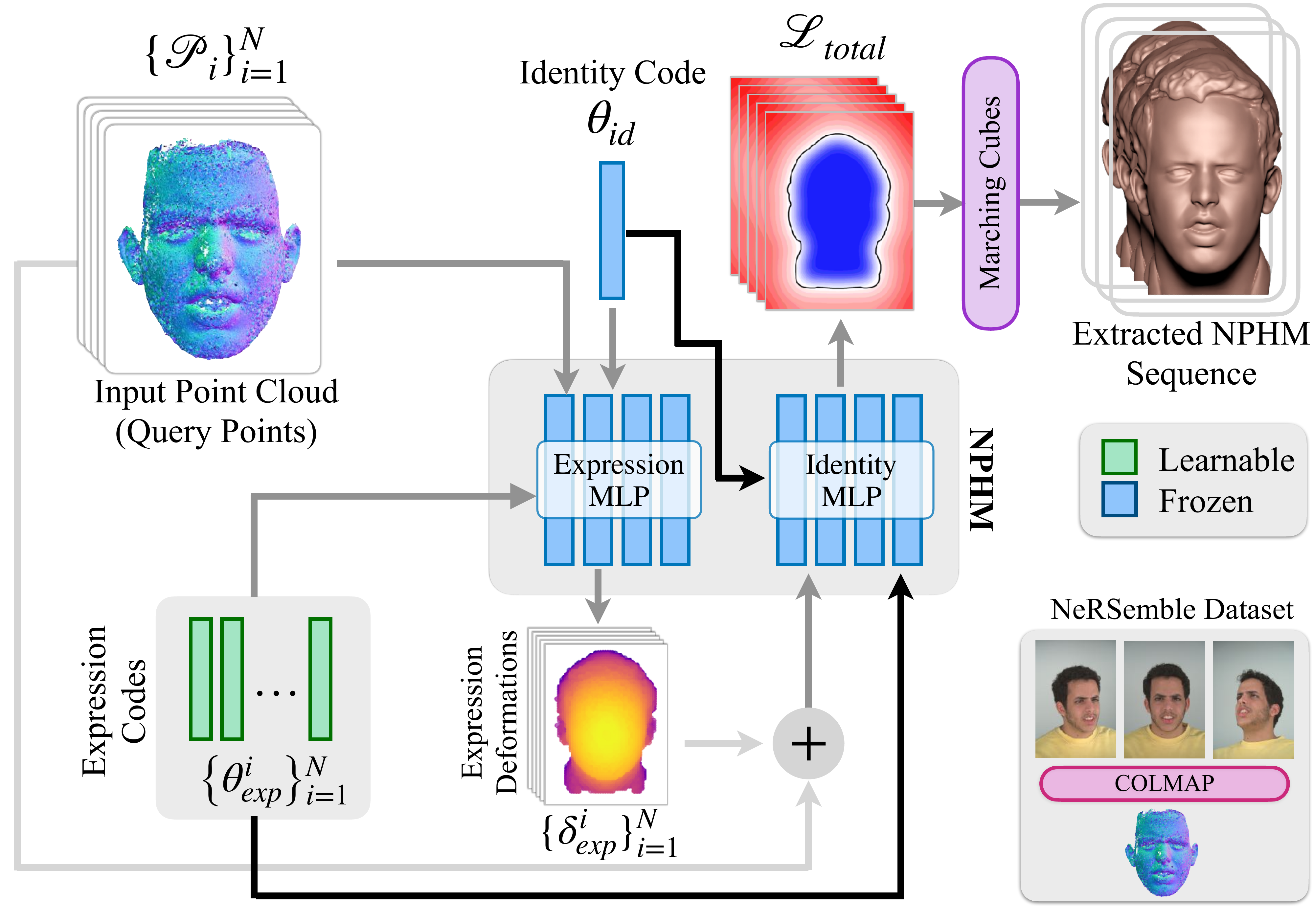}
    \caption{Given the pointcloud sequence $ \big\{ \mathcal{P}_{i} \big\}_{i=1}^{N}$ extracted from multi-view sequences from NeRSemble dataset~\cite{kirschstein2023nersemble} (bottom right), which also act as query points, we leverage the pretrained Expression MLP $\big\{\mathcal{E} \big\}$ to extract the expression deformations $\big\{ \boldsymbol{\delta}^{i}_{exp} \big\}^{1:N}$ and add them back to the input points to get the deformed points $ \big\{ \mathcal{P}^{'}_{i} \big\}_{i=1}^{N}$. These points are then fed to the Identity MLP $\big\{\mathcal{I} \big\}$ which outputs the SDF. The expression codes $ \big\{ \boldsymbol{\theta}_{{exp}}^{i} \big\}^{1:N}$ are optimized using overall loss $\mathcal{L}_{total}$. Note that both fixed identity code $\boldsymbol{\theta}_{{id}}$ and learnable expression codes  $ \big\{ \boldsymbol{\theta}_{{exp}}^{i} \big\}^{1:N}$ are fed to both $\big\{\mathcal{I} \big\}$ and $\big\{\mathcal{E} \big\}$.   Once optimized, the meshes are then extracted with Marching Cubes~\cite{marching_cubes}.} 
    \label{fig:dataset_gen}
\end{figure}

\begin{figure*}[h!]
  \centering
  \includegraphics[width=1.0\linewidth]{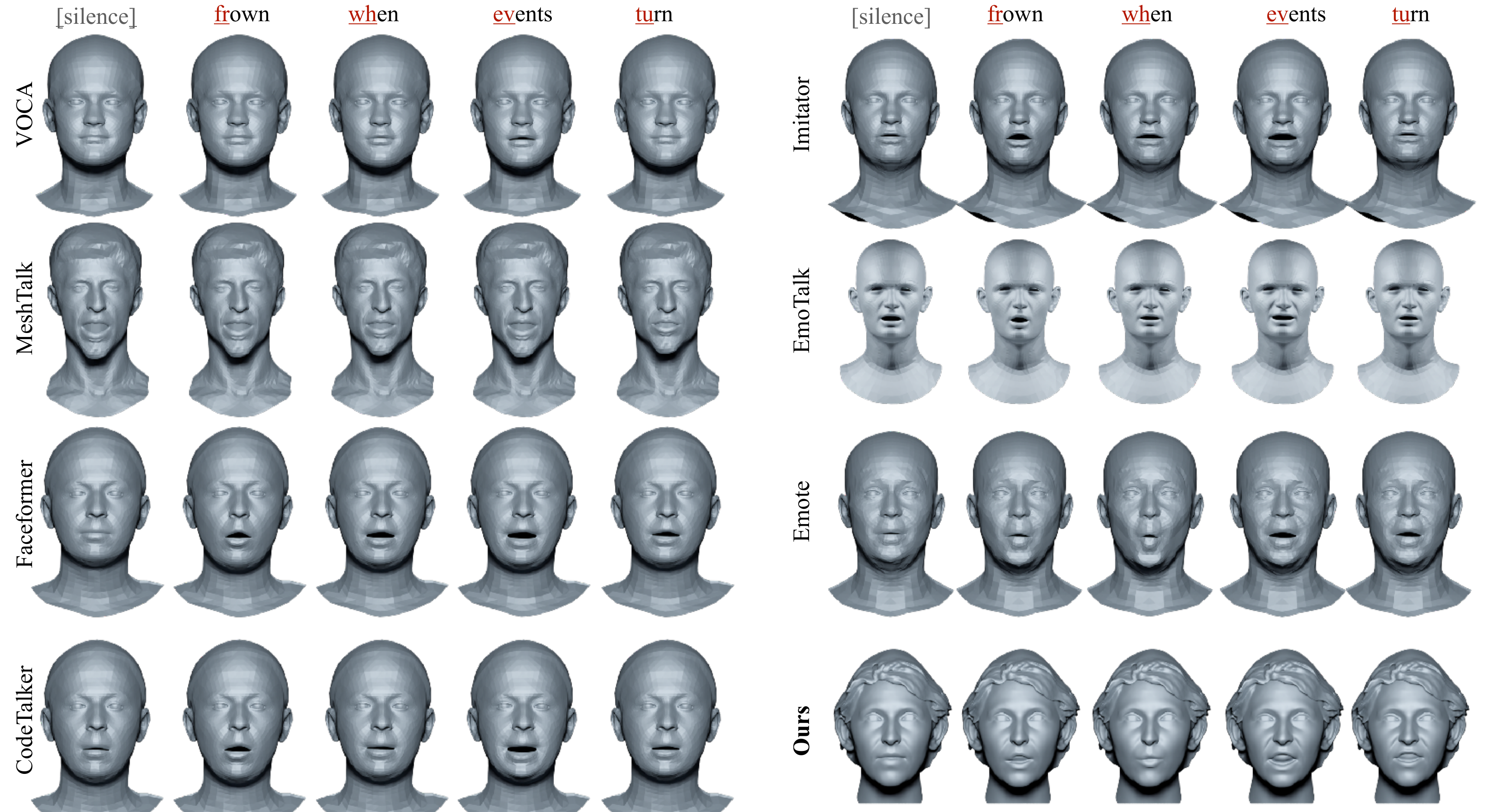}
    \vspace{-0.6cm}
  \caption{Qualitative comparison for audio-driven face animation. 
  Our approach maintains high fidelity while demonstrating rich mouth and nasolabial movements. In particular, we demonstrate more accurate lip articulation, precisely synchronized to phonetic movements.}
  \label{fig:baseline_comp}
\end{figure*}

\section{Dataset Creation}\label{sec:dataset_generation}
To train our latent diffusion model, we require temporally consistent NPHM expression codes as well as their corresponding audio.
To this end, we leverage multi-view recordings of NeRSemble~\cite{kirschstein2023nersemble} captured with 16 cameras of talking human faces paired with corresponding audio signals.
Notably, these sequences are captured for faces already present in the identity space of the NPHM model.
Thus, for a given subject, only the expression codes corresponding to the multi-view sequences have to be estimated.
Specifically, we first extract 3D pointclouds $ \big\{ \mathcal{P}_i \big\}_{i=1}^{N}$ from individual frames using COLMAP~\cite{schoenberger2016sfm},  where $N$ refers to the number of frames in a sequence, with each $ \mathcal{P}_i \in \mathbb{R}^{K \times{3}} $ consisting of $K$ points.
Given the pretrained NPHM model, for the known facial shape $\boldsymbol{\theta_{id}}$, we then optimize the expression codes $ \big\{ \boldsymbol{\theta}^{i}_{\boldsymbol{exp}} \big\}_{i=1}^{N}$ to match the  pointclouds.
In particular, at every iteration, we randomly sample a subset $\mathcal{S}_i$ of 5000 points from $\mathcal{P}_i$ as $\mathcal{S}_i \sim \mathtt{Uniform}(5000, \mathcal{P}_i)$,
and then feed the sampled points $\mathcal{S}_i$, identity code $\boldsymbol{\theta_{id}}$ and learnable expression code $\boldsymbol{\theta}^{i}_{\boldsymbol{exp}}$ into the expression decoder $\big\{\mathcal{E} \big\}$ to obtain the expression deformation $\boldsymbol{\delta}_i$ for the sampled points $\mathcal{S}_i$.
The deformation $\boldsymbol{\delta}_i$ is then added to $\mathcal{S}_i$ to obtain deformed points $\mathcal{D}_i = \mathcal{S}_i + \boldsymbol{\delta}_i$.
These deformed points $\mathcal{D}_i$ are finally fed to the identity decoder $\big\{\mathcal{I} \big\}$ along with latent codes $\boldsymbol{\theta_{id}}$ and $\boldsymbol{\theta}^{i}_{\boldsymbol{exp}}$ to obtain the SDF.
The expression codes are then optimized using the loss $\mathcal{L}_{total}$ which is illustrated in Fig.~\ref{fig:dataset_gen}.
Naively optimizing the expression codes on a per-frame basis shows flickering artifacts.
To mitigate this, we optimize expressions in groups of $n=10$ frames in a sliding window fashion (with an overlap of 2 frames between adjacent windows) using an additional temporal regularization  $\mathcal{L}_{temp}$.
To further prevent the expression codes $ \big\{ \boldsymbol{\theta}^{i}_{\boldsymbol{exp}} \big\}_{i=1}^{N}$ from deviating too much from the distribution already learned by the NPHM model, we employ additional L2 expression regularization $\mathcal{L}_{exp}$.
We minimize the objective $\mathcal{L}_{total}$ defined as:
\begin{equation}
\label{eq:dataset_creation}
    \mathcal{L}_{total} = \lambda_{sdf} \mathcal{L}_{sdf} + \lambda_{temp} \mathcal{L}_{temp} + \lambda_{reg} \mathcal{L}_{reg}. 
\end{equation}
The SDF loss $\mathcal{L}_{sdf}$, temporal regularization $\mathcal{L}_{temp}$, and expression regularization $\mathcal{L}_{exp}$ can be defined as:
\begin{equation}
\mathcal{L}_{sdf} = \sum_{i=1}^{|n|} \frac{1}{|S|}  \sum_{k=1}^{|S|} \big\| \mathtt{SDF} (\mathcal{S}_i^k)   \big\|_1
\end{equation}
\vspace{-0.2cm}
\begin{equation}
 \mathcal{L}_{temp} = \sum_{i=1}^{|n|}  \big\| \boldsymbol{\theta}_{exp}^{i+1} - \boldsymbol{\theta}_{exp}^{i}  \big\| _{\epsilon}
 , ~~~~
 \mathcal{L}_{exp} = \sum_{i=1}^{|n|}  \big\|  \boldsymbol{\theta}_{exp}^{i} \big\|_2,
\end{equation}
where $\|.\|_1$, $\|.\|_2$ and $\|.\|_{\epsilon}$ refers to L1, L2 and Huber loss respectively, and $n$ refers to number of frames simultaneously optimized.
More details regarding aligning the coordinate system of NeRSemble dataset~\cite{kirschstein2023nersemble} with NPHM can be found in the supplemental document.
Once optimized, next to generate mesh sequences, we uniformly sample points from a 3D grid $\mathcal{P} \in \mathbb{R}^{|X| \times |Y| \times |Z|}$ such that $\mathcal{P}_{xyz} =[p_x, p_y, p_z] $ is a point in 3d space.
These points, along with the identity $\boldsymbol{\theta_{id}}$ and expression codes $ \big\{ \boldsymbol{\theta}^{i}_{\boldsymbol{exp}} \big\}_{i=1}^{N} $ are passed to the pretrained NPHM model outputting an SDF; meshes are then extracted with ~\cite{marching_cubes}.
In total, our \OURS{} dataset consists of \dataTotal{} sequences, an order of magnitude larger than the existing datasets~\cite{VOCA2019}.

\begin{figure*}[t!]
  \centering
  \includegraphics[width=1.0\linewidth]{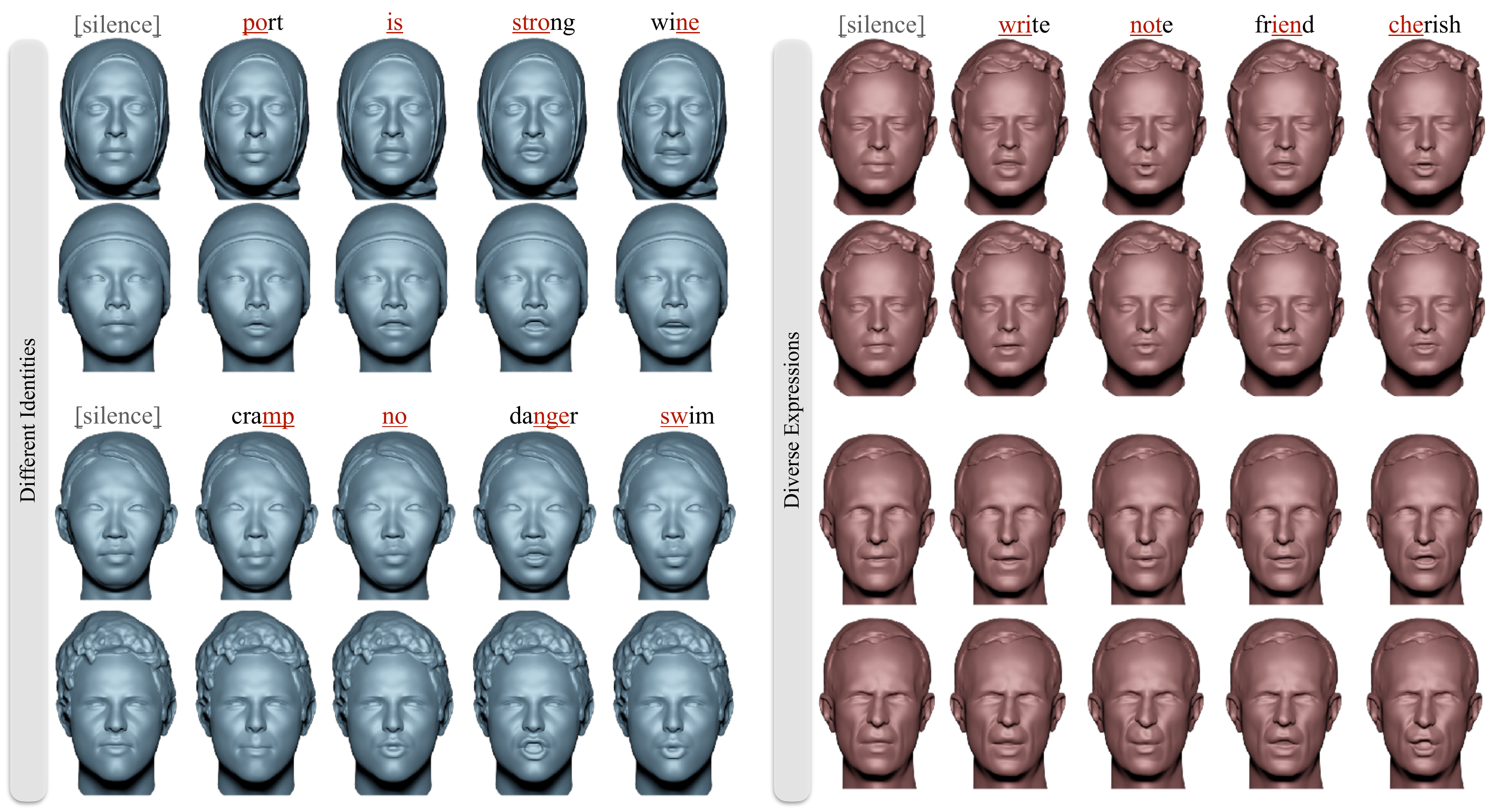}
  \vspace{-0.85cm}
  \caption{\textbf{Left.} Expressions generated by our method can easily be applied to diverse identities with complex geometry. \textbf{Right.} Given a speech signal, our method can further generate diversity in expression for the same identity. Note the difference in speaking style (intensity of mouth opening) as well as eyeblinks/frowning in the upper face area.}
  \label{fig:generative_synthesis}
\end{figure*}

\section{Results}
\label{sec:results}
We evaluate \OURS{} on the task audio-driven motion synthesis and compare it against state-of-the-art methods.

\paragraph{Evaluation Metrics. } Unlike template-based methods that rely on LVE (Lip Vertex Error), our method produces SDFs for each frame; extracted mesh topologies thus vary across frames, making LVE inapplicable for evaluation. Thus, we employ LSE-D (Lip Sync Error Distance)~\cite{wave2lip2020} for quantitative evaluation of lip synchronization. For diversity, we compute FID and KID, as well as diversity scores~\cite{ren2023diffusion}. As FID/KID may not reflect all quality considerations, when samples are limited, we include two established quality metrics: (1) FIQA (Face Image Quality Assessment)~\cite{SDD_FIQA2021} (2) VQA (Video Quality Assessment)~\cite{wu2023dover}. FIQA quantifies image quality for face recognition and similarity to in-the-wild real faces and VQA measures overall video quality in terms of distortions/semantics/aesthetics. Finally, for perceptual evaluation, we perform a user study with 40 participants on a diverse set of 15 unseen audio clips.
Most baseline methods are trained using Vocaset~\cite{VOCA2019}. Thus, for a fair comparison, we construct a hybrid audio test set of 100 sequences, with 25 test audio clips each from Vocaset ($\sim$ 2-3 seconds) and ours ($\sim$ 2-3 seconds), and 50 ($\sim$ 5-7 seconds) from LJSpeech~\cite{ljspeech17}.
Test audios were selected from identities not seen during training of any method. The metrics are reported on this hybrid test set.


\paragraph{Implementation Details.}
For dataset creation, we optimize NPHM expressions for 500 iterations in a group of $n=10$ frames with a step size of 0.001 for iterations $\leq$ 300, and 0.0001 otherwise. We use $\lambda_{sdf}=10$, $\lambda_{temp}=0.1$ and $\lambda_{reg}=0.0025$ for Eq.~\ref{eq:dataset_creation} during optimization. To train our diffusion model, we randomly clip sequences to 2 seconds (48 frames) for efficient minibatching. We train with Adam optimizer with a learning rate of 0.0001. For diffusion, we use 1000 noising/denoising timestamps, a cosine schedule to add noise to input sequences. We apply data augmentation to the expression codes by uniformly modulating them.

\begin{figure}[h!]
  \centering
  \includegraphics[width=0.77\linewidth]{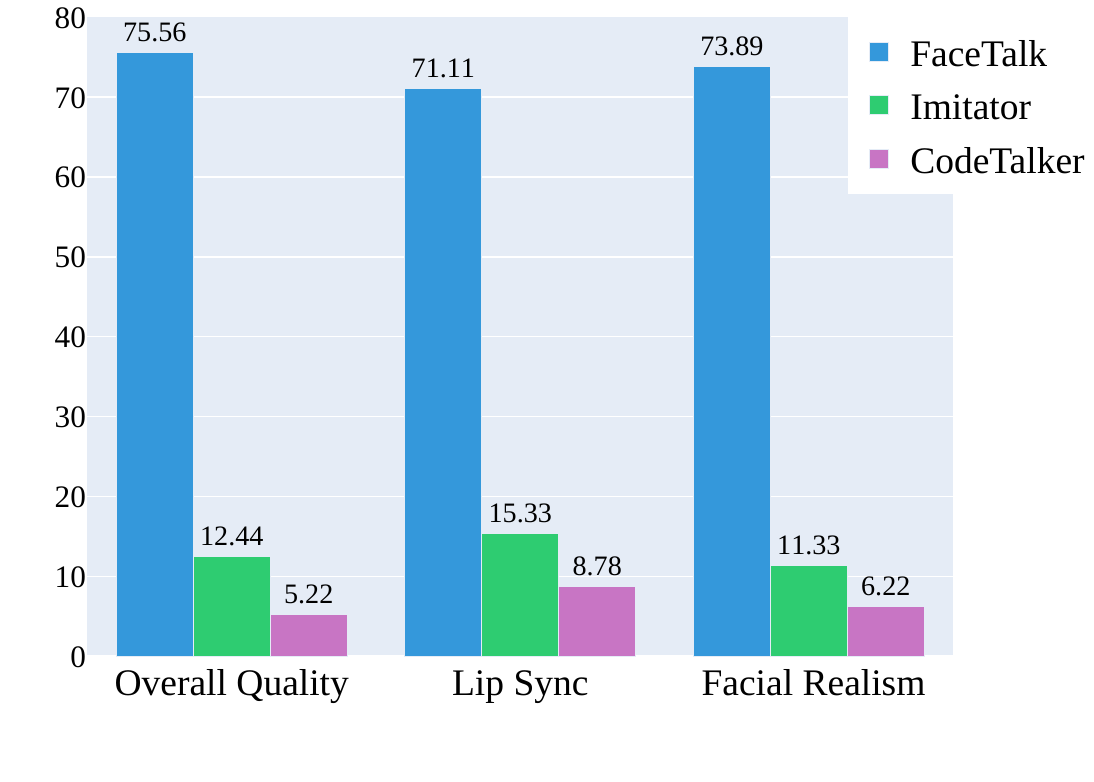}
  \vspace{-0.6cm}
  \caption{User study comparison with baselines. We measure preference for (1) Overall Animation Quality, (2) Lip Synchronization and (3) Facial Realism. \OURS{} results are overwhelmingly preferred over the best baseline methods on all these aspects. }
  \label{fig:user_study}
\end{figure}

\paragraph{Baseline Comparisons.} 
This is the first work to perform speech-conditioned synthesis for volumetric head motion sequences.
We compare with state-of-the-art template-based methods in Fig~\ref{fig:baseline_comp} and Tab.~\ref{tab:baseline_comparison}.
Specifically, we compare against speech-driven animation methods \cite{VOCA2019, richard2021meshtalk, faceformer2022, xing2023codetalker}, personalized methods ~\cite{Thambiraja_2023_ICCV}, as well as recent emotion-based methods~\cite{Peng_2023_ICCV, EMOTE}.
Our approach more accurately captures subtle mouth movements compared to less expressive baselines, resulting in better FID/KID. These scores were evaluated only on mouth region crops to avoid bias towards rest of facial geometry.
Our method consistently achieves better lip-audio synchronization while also representing fine-scale facial details like creasing in the nasolabial folds with wider mouth motions.
This is confirmed by our perceptual user study in Fig~\ref{fig:user_study}.

\begin{table}[t]
    \begin{center}
    \resizebox{1.0\linewidth}{!}{
    \begin{tabular}{l|l|l|l|l|l}
        \toprule
        Method & {LSE-D} $\downarrow$ & FID $\downarrow$   & KID $\downarrow$ & FIQA $\uparrow$ & VQA $\uparrow$   \\
        \toprule

        VOCA~\cite{VOCA2019} & 13.6191 & 239.043 & 0.280 & 33.36 & 0.4251  \\
        MeshTalk~\cite{richard2021meshtalk} & 12.9607 & 220.172 & 0.254 & 36.14 & 0.4855 \\
        FaceFormer~\cite{faceformer2022} & 11.9848 & 215.274 & 0.222 & 38.24 & 0.5227  \\
        CodeTalker~\cite{xing2023codetalker} & 11.8054 & 208.064 & 0.207 & 38.38 & 0.5274   \\
        Imitator~\cite{Thambiraja_2023_ICCV} & 11.6119 & 208.479 & 0.212 & 38.44 & 0.5384   \\
        EmoTalk~\cite{Peng_2023_ICCV} & 11.7485 & 201.311 & 0.201 & 31.80 & 0.5037 \\
        EMOTE~\cite{EMOTE} & 11.7192 & 227.924 & 0.247 & 39.44 & 0.5149 \\
        Ours  & \textbf{11.2737} & \textbf{40.692} & \textbf{0.009} & \textbf{45.75} & \textbf{0.6145}   \\
        \hline
    \end{tabular}
    }
    \end{center}
    \vspace{-0.5cm}
    \caption{In comparison to state of the art, \OURS{} more accurately matches the audio, while producing high perceptual fidelity.}
    \label{tab:baseline_comparison}
\end{table}

\begin{table}[h!]
    \begin{center}
    \resizebox{0.7\linewidth}{!}{
    \begin{tabular}{l|l|l}
        \toprule
        Our Method & {LSE-D} $\downarrow$ & Diversity $\uparrow$  \\
        \toprule
        w/o expr. aug. & {11.6229} & 1.61e-8 \\
        w/o facial smoothing & 11.3488 & 0.34   \\
        w/o FiLM layer & 12.029 & 0.006  \\
        w/o expr.-audio align. & 13.720 & 5.17e-8  \\
        w/o diffusion & 11.3217 & 0.000 \\
        Full (Ours) & \textbf{11.2737} & \textbf{0.34} \\
        \hline
    \end{tabular}
    }
    \end{center}
    \vspace{-0.5cm}
    \caption{Ablation over model design. Expression augmentation improves diversity, while facial smoothing alleviates inter-frame jitter. Using FiLM conditioning achieves accurate mouth pose and increased diversity. Without expression-audio alignment, the model ignores the audio signal. Without diffusion, the model fails to generate diverse results. Our full model with all components achieves the best results. }
    
    \label{tab:model_ablation}
\end{table}

\paragraph{Generative Synthesis.}   Our generated expression codes are identity-agnostic, and are easily transferred to different identities, as shown in Fig.~\ref{fig:generative_synthesis}. Additionally, we can synthesize diverse expressions per-identity for a given audio.

\begin{figure}[h!]
  \centering
  \includegraphics[width=1.0\linewidth]{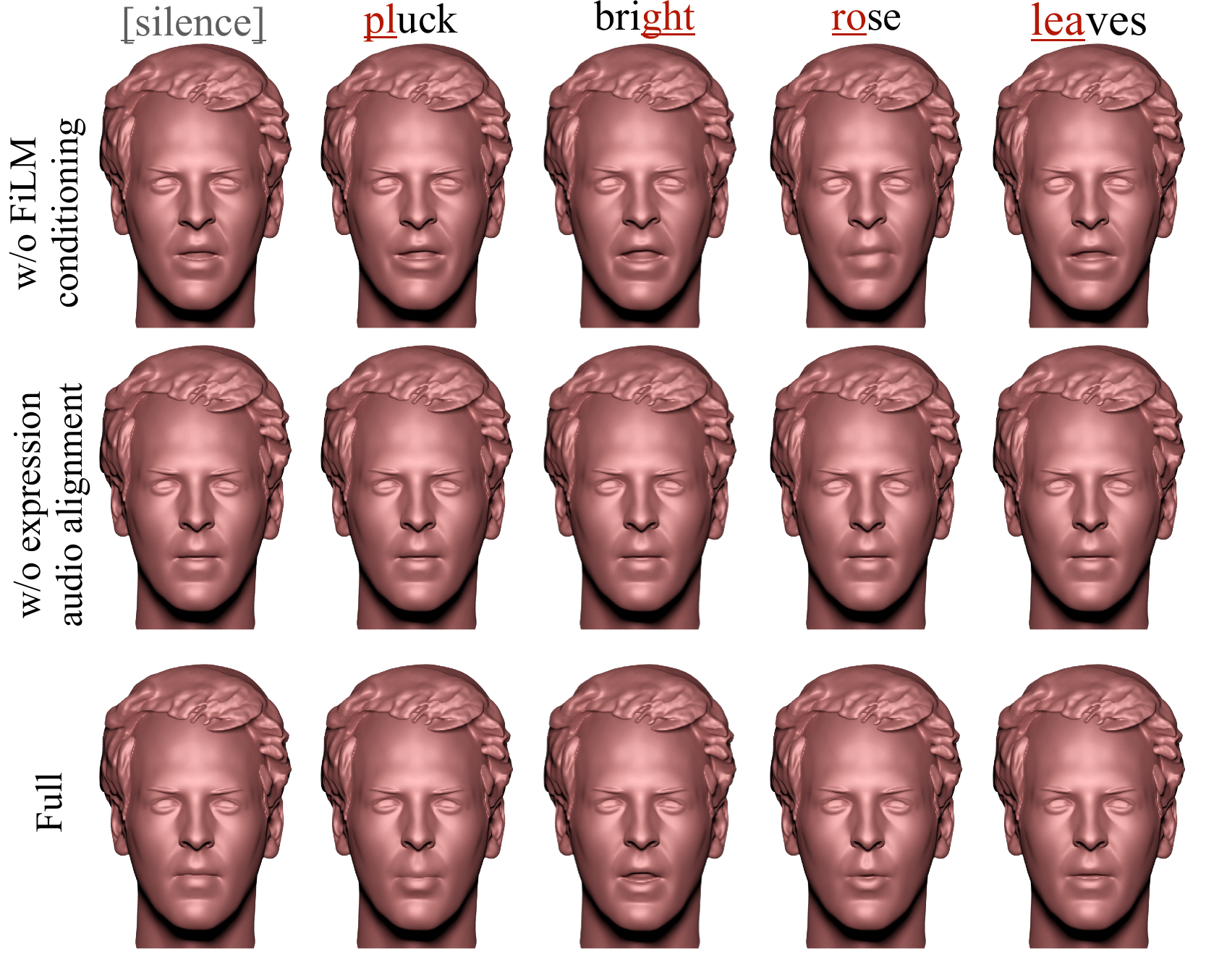}
  \vspace{-0.7cm}
  \caption{Without FiLM conditioning, the model fails to synthesize accurate lip synchronization and even generates uncanny expressions. Without expression-audio alignment, the model fully ignores the audio signal synthesizing constant expression. Our full diffusion model synthesizes accurate lip articulation while maintaining temporal coherence which can be seen in the suppl. video.}
  \label{fig:model_ablation}
\end{figure}

\paragraph{Architecture Ablations.} We ablate our model design choices in Tab~\ref{tab:model_ablation}, with visual results shown in Fig~\ref{fig:model_ablation}, and refer to the supplemental for video demonstration. Our proposed expression augmentation helps the model to synthesize diverse expressions. Facial smoothing removes  unwanted wobbliness in the head and neck region, thereby reducing the inter-frame jitter. FiLM conditioning ensures that expressions generated by our method are more pronounced and that expressions accurately match the audio signal. The expression-audio alignment mask is necessary to match the synthesized expression codes with the audio signal. Our diffusion training ensures that expressions generated by our method are high-quality and diverse. 

\paragraph{Limitations.} While \OURS{} effectively synthesizes high-fidelity and audio-synchronized facial expressions, it still has limitations. For instance, the use of a diffusion model requires multiple denoising steps during inference, limiting its real-time application. We believe that this could be improved by investigating efficient sampling techniques~\cite{watson2021learning}. Currently, our method specializes in synthesizing only the expression codes. For holistic 3D facial animation,  we need to extend its capability to synthesize facial identities. In the future, we would like to generate diverse identities aligned with the gender inferred directly from the audio.

\section{Conclusion}
\label{sec:conclusion}
\vspace{-0.2cm}
In this work, we have introduced \OURS{},  a new approach to synthesize animations of realistic volumetric human heads from audio.
We introduced the first latent diffusion model for audio-driven head animation, producing significantly higher fidelity results than existing methods.
\OURS{} leverages a parametric head model producing volumetric head representations to generate expressive, fine-grained motions such as eye blinks, skin creasing, etc. 
We also demonstrate the applicability of our method for other conditioning signals such as facial landmarks. 
We believe this is an important first step towards enabling the animation of highly detailed 3D face models, which can enable many new possibilities for content creation and digital avatars.

\section{Acknowledgments}
\vspace{-0.2cm}
This work was supported by the ERC Starting Grant Scan2CAD (804724), the Bavarian State Ministry of Science and the Arts and coordinated by the Bavarian Research Institute for Digital Transformation (bidt), the German Research Foundation (DFG) Grant ``Making Machine Learning on Static and Dynamic 3D Data Practical,'' the German Research Foundation (DFG) Research Unit ``Learning and Simulation in Visual Computing,'' and Sony Semiconductor Solutions Corporation. We would like to thank Simon Giebenhain and Tobias Kirschstein for their help with dataset.

\setcounter{section}{0}


\renewcommand\thesection{\Alph{section}}

\section*{Supplemental Material}
We provide additional ablation studies regarding flame baselines, landmark conditioning, dataset creation and model architecture in Section~\ref{sec:additional_ablation}, network architecture details in Section ~\ref{sec:network_architecture}, metric evaluation in Section~\ref{sec:metric_eval}, and additional details regarding dataset creation in Section ~\ref{sec:dataset_creation}. For a visual comparisons of all experiments and ablation studies, we request readers to watch the supplemental video.

\section{Additional Ablation Studies}\label{sec:additional_ablation}

\paragraph{Baselines trained on FLAME fittings of our data.} We report results on our hybrid test set with audio from Vocaset, Ours, and LJSpeech. We additionally train baselines on our dataset (using FLAME fittings, as they don't operate on NPHM space) and show results on the same hybrid test set in Tab.~\ref{tab:baseline_comparison_flame}. Emote and EmoTalk require emotional correspondences and MeshTalk is incompatible with the FLAME topology; thus, a comparison with them is not possible. Our method outperforms baselines on lip-sync (lowest LSE-D), while significantly improving quality (highest FIQA/VQA).

\begin{table}[h]
    \begin{center}
    \resizebox{0.8\linewidth}{!}{
    \begin{tabular}{l|l|l|l}
        \toprule
        Method & LSE-D $\downarrow$ & FIQA $\uparrow$ & VQA $\uparrow$ \\
        \toprule
        FaceFormer & 12.1391 & 38.64 & 0.4206   \\
        CodeTalker & 11.8442 &  39.44 & 0.4218   \\
        Imitator & 11.7402 & 37.83  & 0.4492     \\
        FaceDiffuser & 12.7644 & 37.65 & 0.3826   \\
        Ours (Flame) & 11.6839 & 38.53 & 0.4868 \\
        \hline
        Ours (w/o diffusion) & 11.3217 & 42.81 & 0.5469 \\
        Ours (Full)  & \textbf{11.2737} & \textbf{45.75} & \textbf{0.6145}  \\
        \hline
    \end{tabular}
    }
    \end{center}
    \vspace{-0.6cm}
    
    \caption{\footnotesize  All methods, including baselines were trained on our dataset. The baselines and Ours (Flame) were trained with FLAME fittings, and bottom two rows with NPHM fittings. Our Flame baseline also outperforms other methods in lip sync and video quality. Our full model synthesizes highest-quality animations (highest VQA), well-matches real face qualities (highest FIQA), while performing better lip sync (lowest LSE-D). 
    }
    \label{tab:baseline_comparison_flame}
\end{table}

\begin{figure}[h]
  \centering
  \vspace{-0.2cm}
  \includegraphics[width=1.0\linewidth]{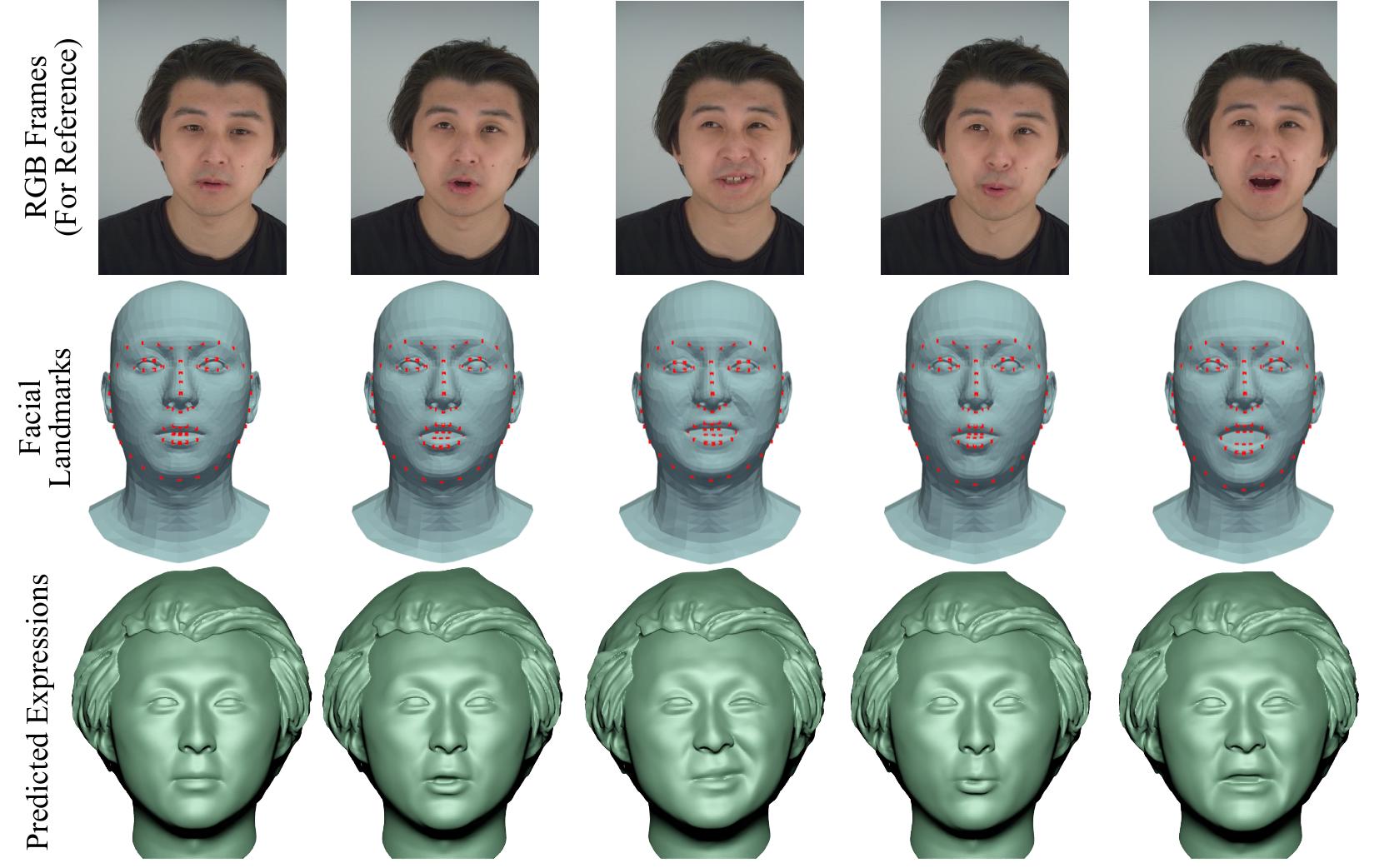}
  \vspace{-0.7cm}
  \caption{Landmark conditioned expression generation. \OURS{} can also generate accurate expressions (bottom) conditioned on face landmarks (middle), matching the original RGB capture (top). }
  \label{fig:landmark_conditioning}
  \vspace{-0.2cm}
\end{figure}

\paragraph{Additional Conditioning.} \OURS{} can also be flexibly adapted to other temporal conditioning signals, such as face landmarks. In Fig~\ref{fig:landmark_conditioning},  we replace input audio with 3D face landmarks extracted by FLAME~\cite{flame2017}, and train a landmark-conditioned model to synthesize faithful expressions. 

\begin{figure}[h!]
  \centering
  \includegraphics[width=1.0\linewidth]{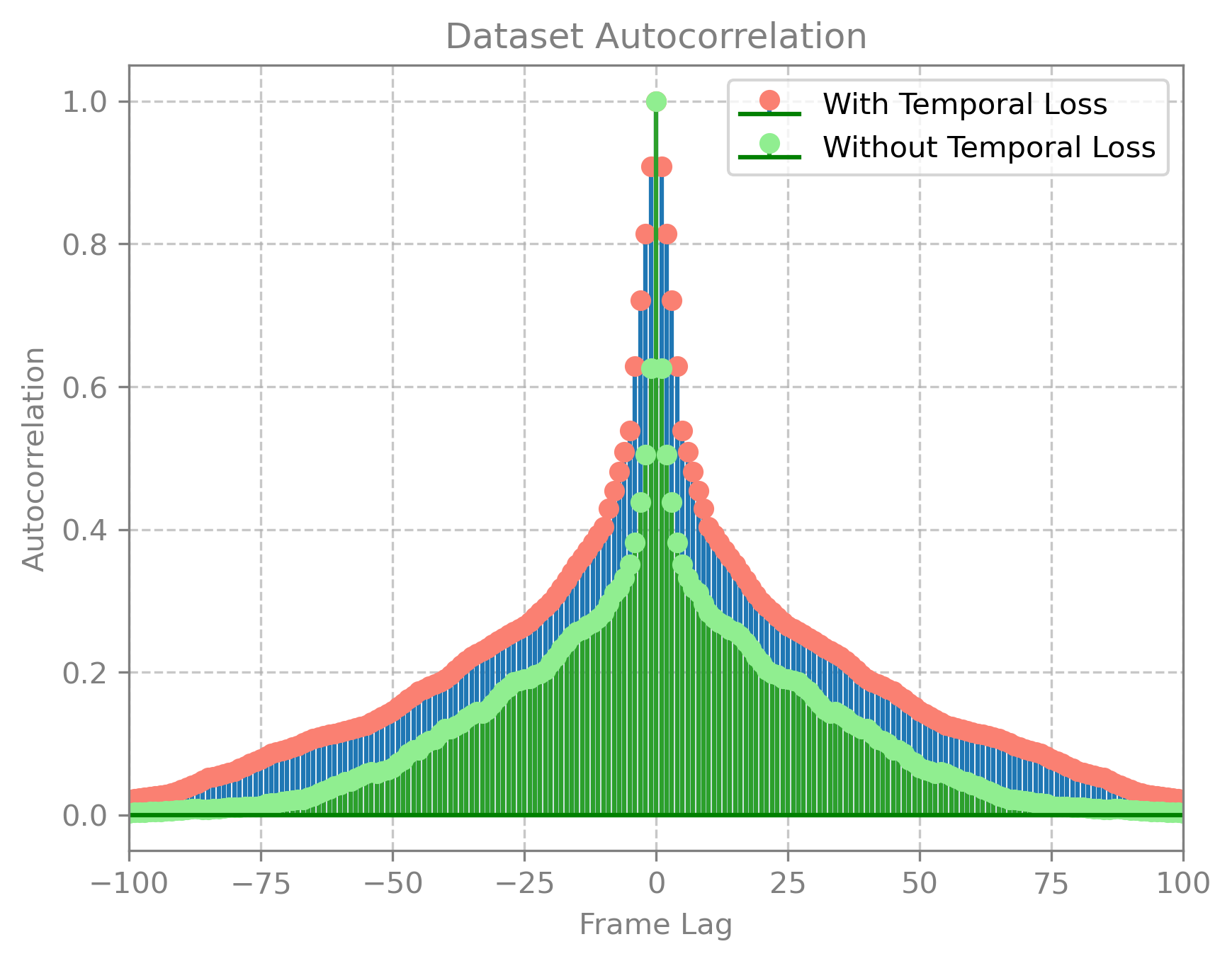}
  \vspace{-0.3cm}
  \caption{Expression Fitting. Note that with temporal loss, the method achieves a high correlation between neighboring frames.}
  \label{fig:exp_fit_acf}
\end{figure}

\paragraph{Expression Fitting.} To evaluate the temporal consistency of our optimized expression sequences, we report Mean Absolute Error (MAE), Root Mean Square Error (RMSE) between adjacent frames and Lip Sync Error Distance(LSE-D) in Tab.~\ref{tab:exp_fit_metrics}. We further visualize auto-correlation between the neighboring frames in Fig~\ref{fig:exp_fit_acf}. We notice that our temporal regularization helps stabilize the high-frequency jitter in the fitted expression codes.

\begin{table}[h!]
    \vspace{0.5cm}
    \begin{center}
    \resizebox{0.8\linewidth}{!}{
    \begin{tabular}{c|l|l|l}
        \toprule
        Method & {MAE} $\downarrow$ & RMSE $\downarrow$ & LSE-D $\downarrow$  \\
        \toprule

        w/o temporal loss & 0.01422 & 0.01978 & 10.9221  \\
        w/ temporal loss & \textbf{0.00248} & \textbf{0.00623} & \textbf{10.5478}  \\
        \hline
    \end{tabular}}
    \end{center}
    \vspace{-0.3cm}
    \caption{Expression Fitting. Our temporal huber loss helps to stabilize the inter-frame jitter while effectively matching the mouth poses with the audio signal. }
    \label{tab:exp_fit_metrics}
\end{table}

\begin{figure*}[h!]
  \centering
  \includegraphics[width=0.7\linewidth]{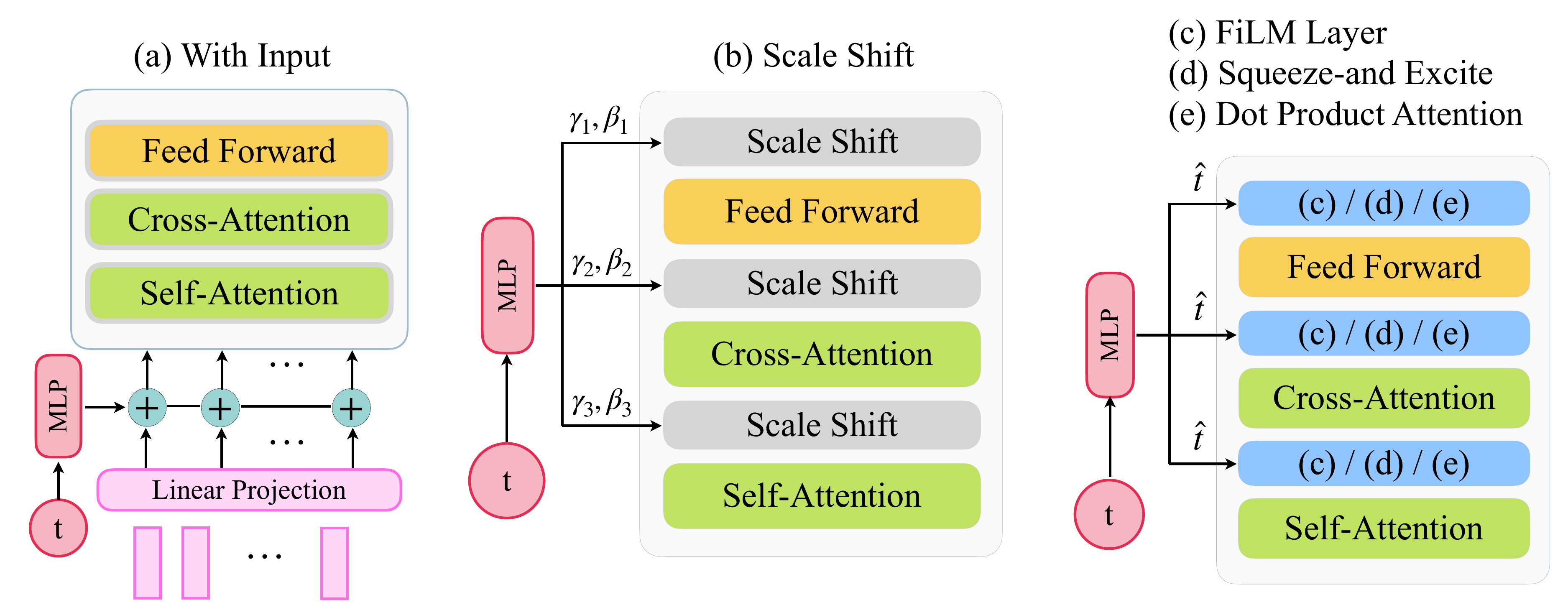}
  \vspace{-0.4cm}
  \caption{Architecture Design Choices to infuse diffusion timestamp into the model. (a)With Input denotes that diffusion timestamp is simply added to the input expression codes. (b) Scale-Shift modulates the intermediate features after Self-Attention (SA), Cross Attention (CA) and FeedForward (FF) layers of the decoder block with diffusion timestamp, via the corresponding scale and shift parameters $\gamma_i, \beta_i$. (c), (d) and (e) inject the diffusion timestamp into the decoder block with one-layer FiLM~\cite{perez2017film}, Squeeze-and-Excite~\cite{hu2019squeezeandexcitation} and vanilla dot-product attention respectively.}
  
  \label{fig:arch_ablations}
\end{figure*}

\paragraph{Model Architecture Ablations.} We ablate different design choices to infuse diffusion timestamps into the model. We experiment with design choices shown in Fig~\ref{fig:arch_ablations} to infuse diffusion timestamps into the model and show results corresponding to these design choices in Tab.~\ref{tab:timestamp_ablation}. We argue that it is critical to carefully inject diffusion timestamps into the model to obtain high-fidelity and diverse results. Using FiLM layers to inject diffusion timestamps outperforms the rest of the design choices, both in producing better lip articulation as well as synthesizing much more diverse results.


\begin{table}[h!]
    \begin{center}
    \resizebox{0.7\linewidth}{!}{
    \begin{tabular}{l|l|l}
        \toprule
        Our Method & {LSE-D} $\downarrow$ & Diversity $\uparrow$\\
        \toprule
        Input & 12.029 & {0.006}  \\
        Scale-Shift & 11.9575 & {0.204}  \\
        SqueezeExcite~\cite{hu2019squeezeandexcitation} & 11.7586 & {0.029} \\
        Dot Prod. Attention & 12.83  & {2.009e-7} \\
        FiLM~\cite{perez2017film} & \textbf{11.2737} & \textbf{0.340}  \\
        \hline
    \end{tabular}}
    \end{center}
    \vspace{-0.25cm}
    \caption{Ablation Study: Effectiveness of different components to fuse diffusion timestamp into the model. Adding it to the input produces muted mouth motion. Using Scale-Shift or Squeeze-Excite layers can perform better lip articulation however produce much less diverse results. Dot Product attention excessively modulates the incoming features, nearly collapsing to a constant expression. Using FiLM layer not only produces better lip articulation but can synthesize much more diverse results.}
    \label{tab:timestamp_ablation}
\end{table}

\begin{figure*}[t!]
  \centering
  \includegraphics[width=0.8\linewidth]{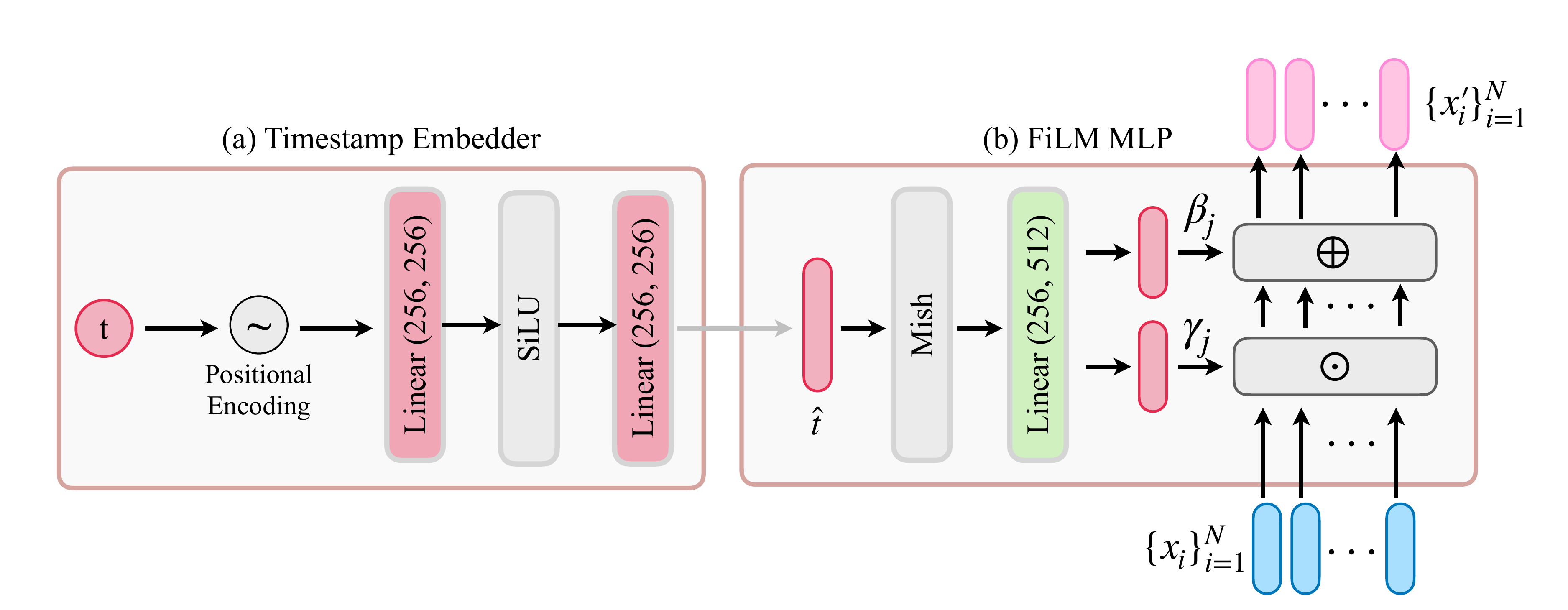}
  \caption{(a) Utilizing a 2-layer MLP architecture with SiLU activation, we project the diffusion timestamp $t$ into the latent space of our expression decoder. The timestamp is first embedded using sinusoidal positional encoding, resulting in a d-dimensional timestamp vector and passed through the MLP layers. (b) FiLM layers are employed to inject the processed diffusion timestamp into the model. The FiLM MLP, featuring Mish activation followed by a linear layer, produces scale $\gamma_j$ and shift $\beta_j$ parameters. These parameters modulate incoming temporal expression features which serves as input for subsequent layers. }
  \label{fig:mlp_architectures}
\end{figure*}

\section{Architecture and Training Details}\label{sec:network_architecture}
\OURS{} is implemented in the Pytorch Lightning framework~\cite{paszke2017automatic, falcon2019pytorch}. For mesh extraction, we use Python-based Marching Cubes library~\cite{pymcubes} and BlenderProc~\cite{Denninger2023} for rendering figures.

\paragraph{Expression Decoder.} Our expression decoder consists of a stack of four transformer-based decoder blocks with FiLM~\cite{perez2017film} layers sandwiched between Multihead Self-Attention, Multihead Cross-Attention and Feedforward Layers, as shown in Fig.~\ref{fig:arch_ablations}(c). We use a stacked multi-head transformer decoder model with latent dimension of 256. For Multihead
Self- and Cross-Attention layers, we use eight heads and set the dimension to 1024 for each transformer decoder block. The linear projection layer at input projects from the NPHM expression space (200-dim.) to the latent space of our expression decoder (256-dim.), and processes them through the stack of decoder blocks and finally use another linear layer to project the learnt expressions from our latent space (256-dim.) back to NPHM expression space (200-dim).

\paragraph{Timestamp Embedder.} To project diffusion timestamp into the latent space of our expression decoder, we use a 2-layer MLP architecture with SiLU~\cite{elfwing2017sigmoidweighted} activation. The diffusion timestamp \( \mathbf{t} \) is first embedded via sinusoidal positional embedding to create d-dimensional timestamp vector as:
\[
\text{{PE}}(t, 2i) = \sin\left(\frac{t}{10000^{2i/d}}\right)
\]

\[
\text{{PE}}(t, 2i+1) = \cos\left(\frac{t}{10000^{2i/d}}\right) ,
\]
where \( t \) is the position and \( i \) is the dimension index. Next, this is passed through the MLP layers and processed timestamp \( \mathbf{\hat{t}} \) is obtained. This is shown in Fig~\ref{fig:mlp_architectures}(a).

\paragraph{FiLM MLP.} We use FiLM layers to inject the processed diffusion timestamp \( \mathbf{t} \) into the the model. The FiLM MLP consists of a Mish activation~\cite{misra2020mish} followed by a linear layer. The linear layer outputs scale $\gamma_j$ and shift $\beta_j$ parameters, which modulates the incoming temporal expression features as:
\begin{equation}
    \big\{ x'_i \big\}_{i=1}^{N} = \bigg( \big\{ x_i \big\}_{i=1}^{N} \odot \gamma_j \bigg) \oplus \beta_j ,
\end{equation}
where $\big\{ x'_i \big\}_{i=1}^{N}$ refers to the processed expression features, which are input to the next layers. This is shown in Fig.~\ref{fig:mlp_architectures}(b).

\section{Metric Evaluation}\label{sec:metric_eval}

\begin{figure}[h!]
  \centering
  \includegraphics[width=0.8\linewidth]{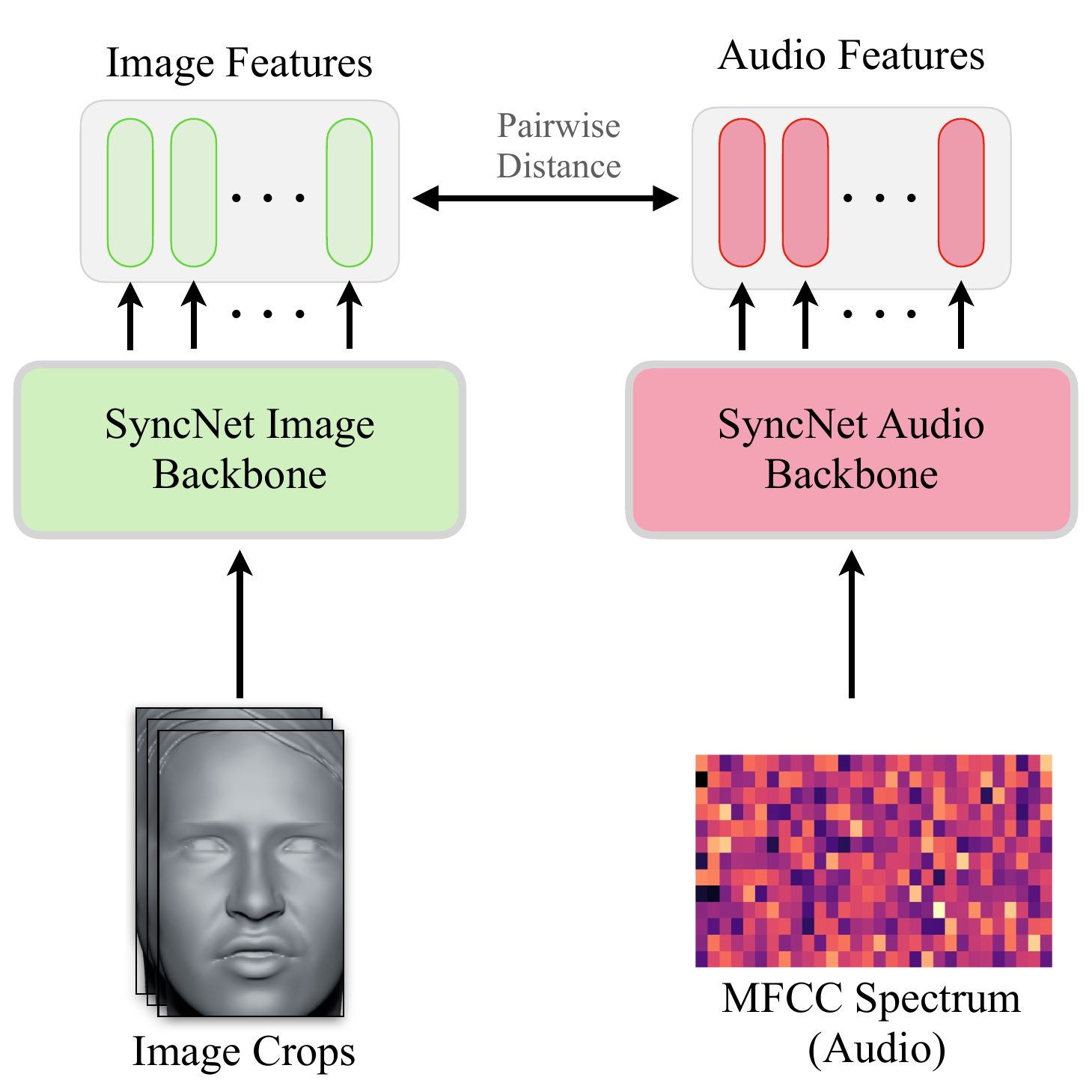}
  \vspace{-0.2cm}
  \caption{The grayscale image crops are passed to the pretrained Syncnet image backbone to extract image features, and audio features, represented as MFCC power spectrum, are extracted via pretrained Syncnet audio backbone. Finally, the pairwise distance between image and audio features are calculated to compute lip synchronization.}
  \label{fig:lsed_d_eval}
\end{figure}

\paragraph{Lip Synchronization} To evaluate lip synchronization of the generated mouth expressions with the audio signal, we use LSE-D (Lip Sync Error Distance)~\cite{wave2lip2020}. Specifically, this involves feeding rendered grayscale face crops and the corresponding audio signal into a pre-trained SyncNet~\cite{Chung16a} to evaluate how close the acoustic signal matches the phonetic movements. The facial movements are encoded as grayscale crops of only the facial region, and the audio signal is represented as MFCC power spectrum. These are then passed into the pretrained SyncNet backbone~\cite{Chung16a} and the pairwise distance is evaluated, as shown in Fig.~\ref{fig:lsed_d_eval}.

\paragraph{Diversity.} To calculate the visual fidelity, we report the standard GAN metrics KID and FID on the grayscale mouth region crops for all the methods, however, it cannot fully capture how diverse the generated expressions are for a given audio signal. We, thus, evaluate the diversity score $D$ in the latent expression space of NPHM model. For the diversity, we measure how much the generated expression codes diversify for the same audio input. Given a set ${S}=\{ S_1, S_2, ... S_K \}$ of generated expression codes (generated from different random noises) for the same audio signal, we calculate the pairwise distance within each set over the entire test dataset as:
\begin{equation}
    D = \frac{1}{|N| \times |K|} \sum_{i=1}^N \sum_{j=1}^K \sum_{l=1, l \neq j }^K  \| \textbf{e}_{i,j} - \textbf{e}_{i,l}  \|_2 ,
\end{equation}
where $|N|$ refers to the number of audio signals in the test set. $|K|$ refers to the number of generated expression codes for the same audio signal, and $\textbf{e}$ refers to the synthesized expression codes.

\begin{figure*}[t!]
  \centering
  \includegraphics[width=1.0\linewidth]{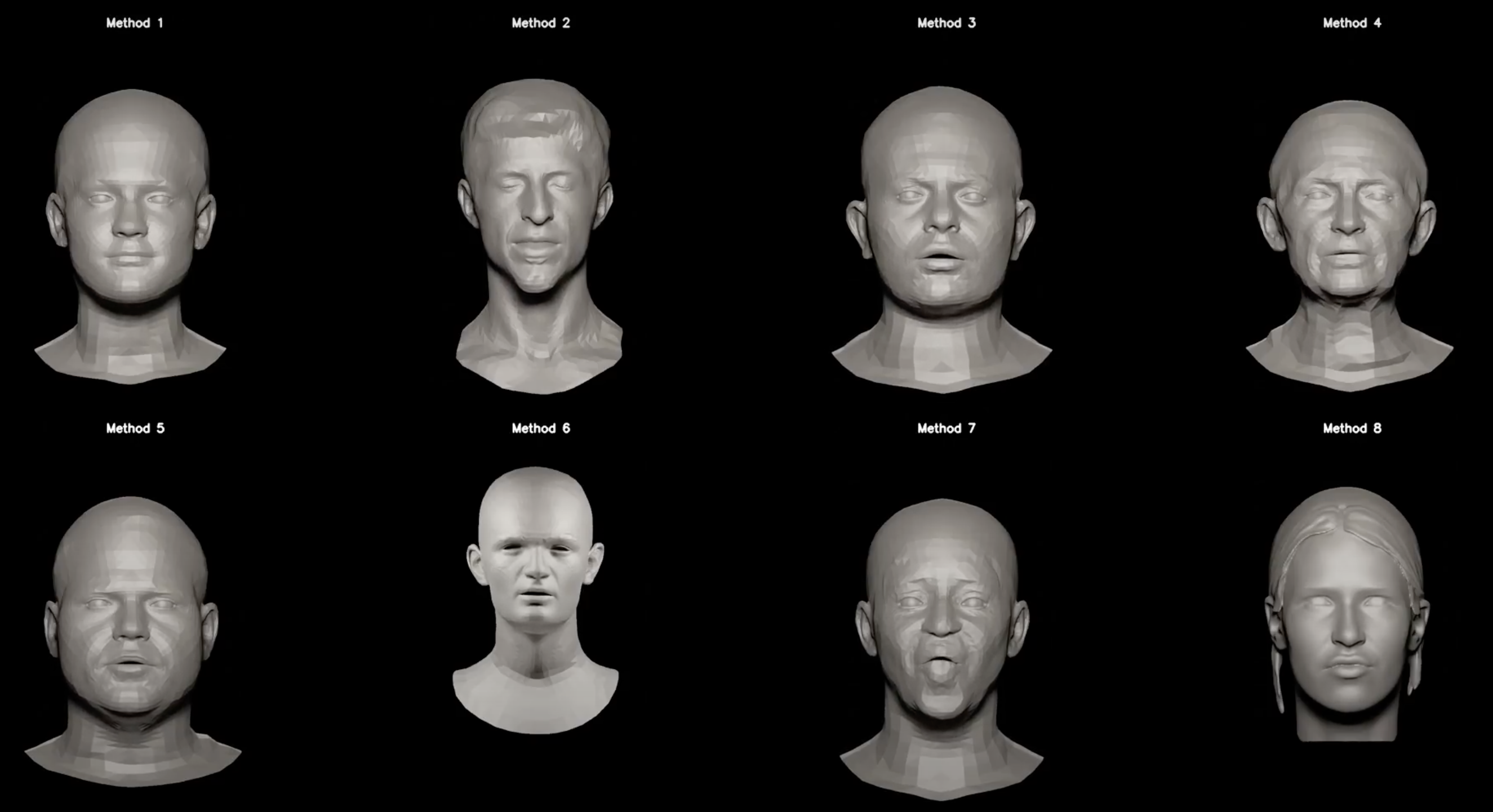}
  \vspace{-0.5cm}
  \caption{Different methods shown to the users during perceptual evaluation. Method names were anonymized to avoid bias towards a particular method. }
  \label{fig:user_study_vid}
\end{figure*}

\begin{figure}[h!]
\vspace{0.5cm}
  \centering
  \includegraphics[width=1.0\linewidth]{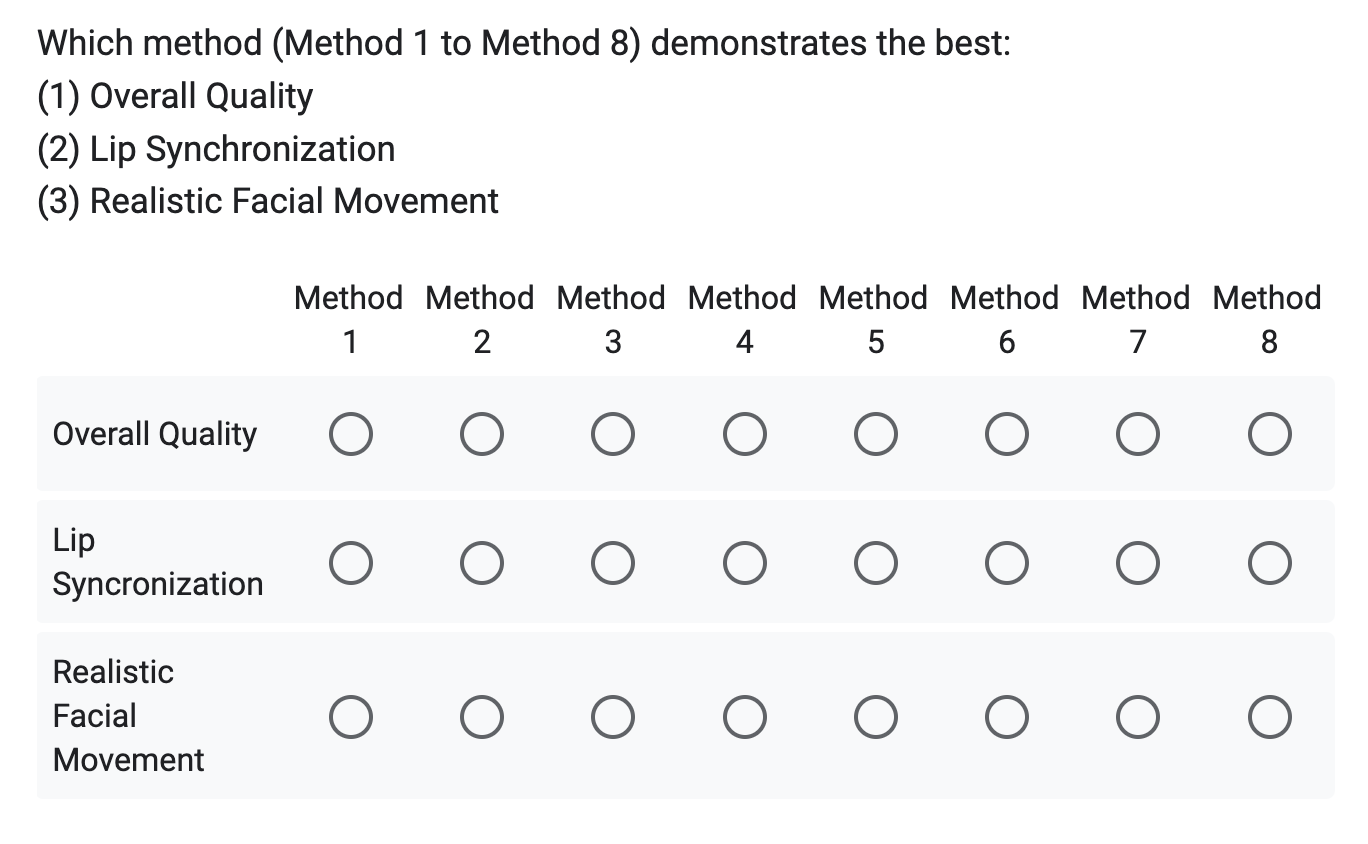}
  \vspace{-1cm}
  \caption{Questions asked during the user study to assess the quality of different methods. Users were asked to select one of the eight methods for each of the three questions based on which one they believe exhibits best results.}
  \label{fig:user_study_questions}
\end{figure}

\paragraph{User Study.} To evaluate the fidelity based on human perceptual evaluation, we performed a user study with 40 participants. The users were given a carefully crafted set of instructions to evaluate (a) Overall Animation Quality (b) Lip Synchronization and (c) Realism in Facial Movements. The users were asked to assess eight different anonymous methods (including \OURS{}), shown in Fig~\ref{fig:user_study_vid} on these three parameters.

In the course of the study, participants were presented with these questions to focus on different aspects of 3D facial animation, shown in Fig~\ref{fig:user_study_questions}. For every question, participants were instructed to meticulously evaluate the provided methods and select the option that best aligned with their judgment.
For the first question evaluating overall quality, participants were instructed to consider factors such as visual appeal, clarity, and general impression, and to choose the method number that they believed demonstrates the highest overall quality.
Moving on to the second question, participants were directed to evaluate the lip synchronization of each 3D facial animation method. They were prompted to pay close attention to how well the lip movements aligned with the spoken words or sounds. Participants were reminded to select only one option that, in their judgment, exhibited the best lip synchronization.
Lastly, the third question honed in on evaluating the realistic facial movement of each 3D facial animation method. Participants were instructed to consider the naturalness and persuasiveness of facial expressions and movements and to choose the method number that, in their opinion, demonstrates the most realistic facial movement. Again, participants were reminded to select only one option per question throughout the study.

\section{Dataset Creation}\label{sec:dataset_creation}
In the work, we leverage the Nersemble dataset~\cite{kirschstein2023nersemble} to create the paired audio-NPHM expression dataset. The Nersemble dataset consists of multi-view recordings of people speaking with corresponding audio, for identities present in the shape space of NPHM model. The full dataset creation process is explained in the following steps. 

\begin{figure*}[t!]
  \centering
  \includegraphics[width=1.0\linewidth]{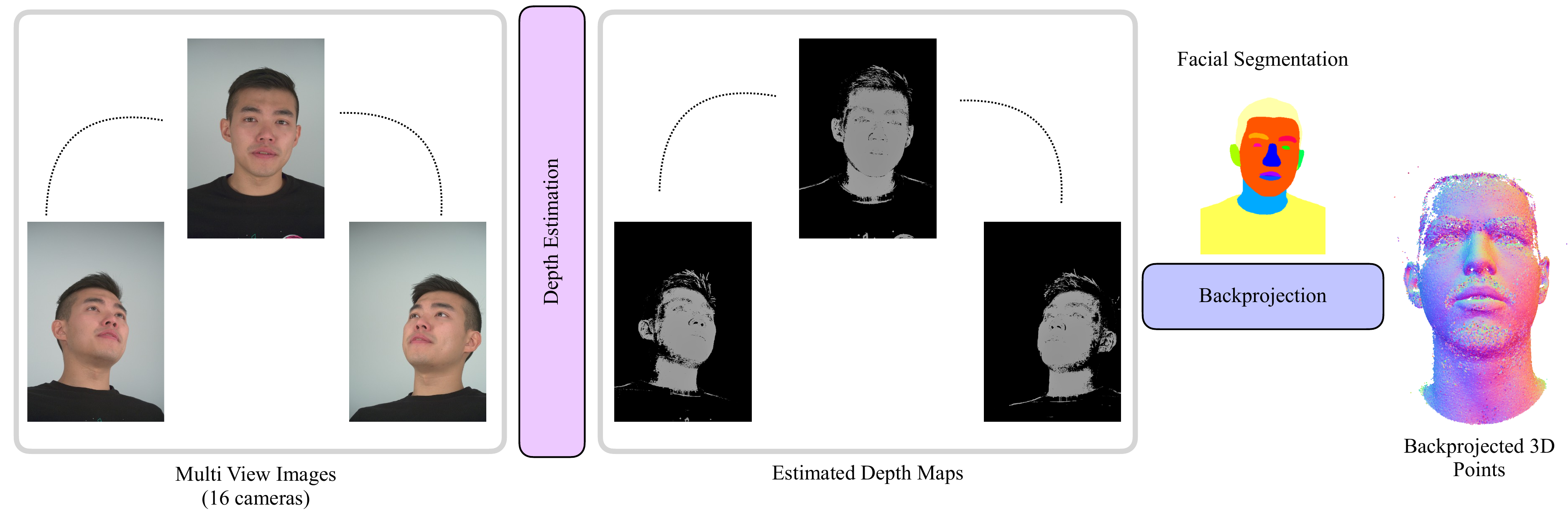}
  \vspace{-0.2cm}
  \caption{Given a set of multi-view images, we first use COLMAP to estimate depth maps, and backproject these to 3D space only for the pixels corresponding to facial area by leveraging facial segmentation mask. These steps are performed for all the frames in the sequence. }
  \label{fig:backprojection}
\end{figure*}

\paragraph{Backprojection.} Given the multi-view frames and camera calibrations from Nersemble dataset, using the estimated depth and normals, we first backproject them to 3D to obtain 3D points and 3D landmarks. A depth mask based on valid depth values (between 0 and 1.4) is used during the process. To include only the valid pixels from the estimated depth map, we use 2D facial segmentation masks to include points only for the facial area. The valid depth values and corresponding screen coordinates are extracted based on the depth mask. Screen coordinates are converted to canonical camera coordinates and then to camera coordinates. The resulting camera coordinates are then transformed to world coordinates using the extrinsic parameters. Normal vectors are extracted based on the depth mask and transformed to world coordinates using the rotation part of the extrinsic parameters. This is shown in Fig~\ref{fig:backprojection}.

\paragraph{Flame Fitting.} Next, we align the Flame template mesh~\cite{flame2017} to the 3D landmarks obtained above from multi-view data. We compute the rigid transformation $[R_{rigid}, t_{rigid}]$ between the FLAME template landmarks and the valid multi-view landmarks. Outliers are filtered based on the Euclidean distance between the transformed FLAME template landmarks and the original multi-view landmarks. A threshold of 0.020 is used to determine valid points. After filtering outliers, a similarity transformation is recomputed to obtain the transformation scale factor, rotation matrix, and translation vector as $[s_{sim}, R_{sim}, t_{sim}]$.

Initializing Flame parameters $\mathcal{F} = [s, R, t, \theta_{shape}, \theta_{exp}]$ with $[s_{sim}, R_{sim}, t_{sim}]$ obtained above and zero vectors for shape $\theta_{shape}$ and expression blendshapes $\theta_{exp}$, we optimize flame parameters $\mathcal{F}$ for entire sequence using an Adam optimizer for 2500 steps, with overall loss $\mathcal{L}_{total}$.

We use several geometric and temporal regularizers to obtain accurate flame fittings. Specifically, $\mathcal{L}_{lmk}$ measures residuals between predicted flame landmarks $L_{flame} \in \mathbb{R}^{68 \times 3}$ and input landmarks from multi-view stereo $L_{mv} \in \mathbb{R}^{68 \times 3}$, normalized over all frames $N$ of the sequence as:
\begin{equation}
    \mathcal{L}_{lmk} = \frac{1}{N} \sum_{i=1}^{N} \big\| L_{flame}^i -  L_{mv}^i \big\|_1
\end{equation}
We use different weights for different regions (jaw, eye, mouth). We also use geometric loss $\mathcal{L}_{geo}$ between predicted Flame vertices $V_{flame} \in \mathbb{R}^{5023 \times 3}$ and backprojected 3D points $P_{mv} \in \mathbb{R}^{K \times 3}$, considering both point-to-point and point-to-plane distances. The point-to-point distance is defined as:
\begin{equation}
    \mathcal{L}_{point} =  \frac{1}{N} \sum_{i=1}^{N} \big\| V_{flame}^i -  P_{mv_{match}}^i \big\|_1,
\end{equation}
and point-to-plane distance is defined as:
\begin{equation}
    \mathcal{L}_{plane} =  \frac{1}{N} \sum_{i=1}^{N} \big\| V_{flame}^i -  P_{mv_{match}}^i \cdot N_{mv_{match}}^i   \big\|_1,
\end{equation}
where $V_{flame}^i \in \mathbb{R}^{5023 \times 3}$ refers to the predicted Flame vertices for the $i^{th}$ frame, $P_{mv_{match}}^i \in \mathbb{R}^{5023 \times 3}$ and $N_{mv_{match}}^i \in \mathbb{R}^{5023 \times 3} $ refer to the points and corresponding normals closest to $V_{flame}^i$. The geometric loss $\mathcal{L}_{geo}$ is defined as:
\begin{equation}
    \mathcal{L}_{geo} = 0.1 . \mathcal{L}_{point} + 0.9 . \mathcal{L}_{plane}.
\end{equation}
Next, we employ parameter regularization to penalize large values of shape, expression, and rigid transformation parameters.
\begin{equation}
    \mathcal{L}_{shape}^{reg} =   \| \theta_{shape} \|_2  
\end{equation}
\begin{equation}
    \mathcal{L}_{exp}^{reg} =   \frac{1}{N} \sum_{i=1}^{N} \| \theta_{exp}^i \|_2  
\end{equation}
\begin{equation}
    \mathcal{L}_{rigid}^{reg} =   \frac{1}{N} \sum_{i=1}^{N} \Bigg( \frac{\| R^i \|_2}{2 \pi} + \| t^i \|_2 + \| s^i \|_2  \Bigg) .
\end{equation}
The overall regularization $\mathcal{L}_{reg}$ is then defined as:
\begin{equation}
    \mathcal{L}_{reg} = \mathcal{L}_{shape}^{reg} + \mathcal{L}_{exp}^{reg}  + \mathcal{L}_{rigid}^{reg}  .
\end{equation}
Finally, we leverage smoothness loss $\mathcal{L}_{smooth}$ to penalize changes in expression and rigid transformations between consecutive frames:
\begin{equation}
    \mathcal{L}_{exp}^{smooth} =   \frac{1}{N} \sum_{i=2}^{N} \| \theta_{exp}^i - \theta_{exp}^{i-1} \|_2  
\end{equation}
\begin{equation}
    \mathcal{L}_{rigid}^{smooth} =   \frac{1}{N} \sum_{i=2}^{N} \Bigg( \frac{\| R^i - R^{i-1} \|_2}{2 \pi} + \| t^i - t^{i-1} \|_2  \Bigg) .
\end{equation}
The total optimization loss is then defined as:
\begin{align}
    \mathcal{L}_{\text{total}} &= \lambda_{lmk}\mathcal{L}_{lmk} + \lambda_{geo}\mathcal{L}_{geo}
     + \lambda_{reg}\mathcal{L}_{reg} \nonumber \\
    &\quad + \lambda_{smooth}\mathcal{L}_{smooth} .
\end{align}

\paragraph{Audio Processing} The audio captured by the Nersemble dataset contains a lot of background noise and speaking volume of the person is very low. We first amplify the audio signal by increasing it by 20 dB (deciBels), however this adds a lot of white noise to the audio signal. We then process the audio signal to remove this added noise by using the NoiseReduce library~\cite{tim_sainburg_2019_3243139}.


{\small
\bibliographystyle{ieeenat_fullname}
\bibliography{egbib}
}

\end{document}